\documentclass[preprint]{elsarticle}

\journal{}

\usepackage[utf8]{inputenc} 	
\usepackage{graphicx}      		
\usepackage[numbers]{natbib}	
\usepackage{amsfonts}        	
\usepackage{amsmath,amssymb}    
\usepackage{bm}					
\usepackage{xcolor}             
\usepackage{amsthm}  
\usepackage{algorithm,algpseudocode}  
\usepackage{tabularx} 
\usepackage{booktabs} 
\usepackage[caption=false]{subfig} 

\newcommand{\reals}{\mathbb{R}}

\newcommand{\expectationp}[2][]{\mathbf{E} \ifx #1 \undefined \else _{#1} \fi \left[#2\right]}
\newcommand{\variancep}[2][]{\mathbf{V} \ifx #1 \undefined \else _{#1} \fi \left[#2\right]}
\newcommand{\dataset}{\mathcal{D}}
\newcommand{\minibatch}{\mathcal{B}}
\newcommand{\kullb}[2]{D_{\text{KL}} \left( {#1} \, \lVert \, {#2} \right)}
\newcommand{\elbo}[1]{\mathcal{L} \left( {#1} \right)}
\newcommand{\condbar}{\,|\,}  
\newcommand{\indep}{\perp \!\!\! \perp}

\newcommand{\smallplus}{\texttt{+}}

\newcommand{\normaldist}{\mathcal{N}}

\begin{document}

\begin{frontmatter}

\title{Bayesian Neural Networks for Virtual Flow Metering: An Empirical Study}

\author[ssaddress,ntnuaddress]{Bjarne Grimstad\corref{mycorrespondingauthor}}
\cortext[mycorrespondingauthor]{Corresponding author}
\ead{bjarne.grimstad@gmail.com}

\author[ssaddress,ntnuaddress]{Mathilde Hotvedt}
\author[ssaddress,uioaddress]{Anders T. Sandnes}
\author[uioaddress]{Odd Kolbjørnsen}
\author[ntnuaddress]{Lars S. Imsland}





\address[ssaddress]{Solution Seeker AS, Oslo, Norway}
\address[ntnuaddress]{Department of Engineering Cybernetics, Norwegian University of Science and Technology, Trondheim, Norway}
\address[uioaddress]{Department of Mathematics, University of Oslo, Oslo, Norway}

\begin{abstract} 
Recent works have presented promising results from the application of machine learning (ML) to the modeling of flow rates in oil and gas wells. Encouraging results and advantageous properties of ML models, such as computationally cheap evaluation and ease of calibration to new data, have sparked optimism for the development of data-driven virtual flow meters (VFMs). Data-driven VFMs are developed in the small data regime, where it is important to question the uncertainty and robustness of models. The modeling of uncertainty may help to build trust in models, which is a prerequisite for industrial applications. The contribution of this paper is the introduction of a probabilistic VFM based on Bayesian neural networks. Uncertainty in the model and measurements is described, and the paper shows how to perform approximate Bayesian inference using variational inference. The method is studied by modeling on a large and heterogeneous dataset, consisting of 60 wells across five different oil and gas assets. The predictive performance is analyzed on historical and future test data, where an average error of 4-6\% and 8-13\% is achieved for the 50\% best performing models, respectively. Variational inference appears to provide more robust predictions than the reference approach on future data. Prediction performance and uncertainty calibration is explored in detail and discussed in light of four data challenges. The findings motivate the development of alternative strategies to improve the robustness of data-driven VFMs.
\end{abstract}


\end{frontmatter}


\section{Introduction}\label{sec:introduction}
Knowledge of multiphase flow rates is essential to efficiently operate a petroleum production asset. Measured or predicted flow rates provide situational awareness and flow assurance, enable production optimization, and improve reservoir management and planning. However, multiphase flow rates are challenging to obtain with great accuracy due to uncertain subsurface fluid properties and complex multiphase flow dynamics \cite{NodalAnalysis2015}. In most production systems, flow rates are measured using well testing. While these measurements are of high accuracy, they are intermittent and infrequent \cite{Monteiro2020}. Some production systems have multiphase flow meters (MPFMs) installed at strategic locations to continuously measure flow rates. Yet, these devices are expensive, and typically have lower accuracy than well testing. An alternative approach is to compute flow rates using virtual flow metering (VFM). VFM is a soft-sensing technology that infers the flow rates in the production system using mathematical models and ancillary measurements \cite{Toskey2012}. Many fields today use some form of VFM technology complementary to flow rate measurements. There are two main applications of a VFM: i) real-time prediction of flow rates, and ii) prediction of historical flow rates. The second application is relevant to the prediction of missing measurements due to sensor failure or lacking measurements in between well tests.

VFMs are commonly labeled based on their use of either mechanistic or data-driven models \cite{Bikmukhametov2020a}. Both model types can be either dynamic or steady-state models. Mechanistic VFM models are derived from prior knowledge about the internal structure of the process \cite{Solle2016}. Physical, first-principle laws such as mass, energy, and momentum-balance equations, along with empirical closure relations, are utilized to describe the relationship between the system variables. Mechanistic modeling is the most common approach in today's industry and some commercial VFMs are Prosper, ValiPerformance, LedaFlow, FlowManager, and Olga \cite{Amin2015}. 

In contrary to mechanistic models, data-driven models exploit patterns in process data and attempt to find relationships between the system variables with generic mathematical models. In other words, data-driven models attempt to model the process without utilizing explicit prior knowledge \cite{Solle2016}. In recent years, there has been an increasing number of publications on data-driven VFMs \cite{Bikmukhametov2020a}. The developments are motivated by the increasing amount of sensor data due to improved instrumentation of petroleum fields, better data availability, more computing power, better machine learning tools and more practitioners \cite{Balaji2018}. Additionally, data-driven VFMs may require less maintenance than a mechanistic VFMs \cite{AlQutami2017c}. Even so, commercial data-driven VFMs are rare. This is arguably due to the following data challenges, which must be overcome by data-driven VFMs:
\begin{enumerate}
    \item Low data volume
    \item Low data variety
    \item Poor measurement quality
    \item Non-stationarity of the underlying process
\end{enumerate}

The first two challenges are due to data-driven methods, especially neural networks, being data-hungry, and require substantial data volume and variety to achieve high accuracy \cite{Mishra2018}. Petroleum production data do not generally fulfill these requirements. For petroleum fields without continuous monitoring of the flow rates, new data is obtained at most 1-2 times per month during well testing \cite{Monteiro2020}, yielding low data volume. For fields with continuous measurements, the data volume may be higher, yet, the second challenge of low variety remains. Low data variety relates to the way production systems are operated and how it affects the information content in historical production data. The production from a well is often kept fairly constant by the operator, in particular during plateau production, i.e., when the production rate is limited by surface conditions such as the processing capacity. When a field later enters the phase of production decline, the operator compensates for falling pressures and production rates by gradually opening the production choke valves. This can introduce correlations among the measured variables which are unfortunate for data-driven models. A common consequence of modeling in the small data regime is overfitting which decreases the generalization ability of the model, that is, the models struggle with extrapolation to unseen operating conditions \cite{Solle2016}. Nonetheless, one should be able to model the dominant behavior of the well and make meaningful predictions close to the observed data if care is taken to prevent overfitting \cite{AlQutami2018}. 

The third challenge, poor measurement quality, highly influences the predictive abilities of data-driven VFMs. Common issues with measurement devices in petroleum wells include measurement noise, bias, and drift. Additionally, equipment or communication failures may lead to temporarily or permanently missing data. Common practices to improve data quality include device calibration, data preprocessing and data reconciliation \cite{Camara2017}. In model development, methods such as parameter regularization and model selection techniques prevent overfitting of the model in the presence of noisy data. However, some of the above issues and practices may be challenging to handle in a data-driven model.

Lastly, the underlying process in petroleum production systems, the reservoir, is non-stationary. The pressure in the reservoir decreases as the reservoir is drained and the composition of the produced fluid changes with time \cite{Foss2018}. Time-varying boundary conditions make it more difficult to predict future process behavior for data-driven VFMs as they often struggle with extrapolation. As mentioned above, methods to prevent overfitting to the training data in model development may (and should) be utilized to improve extrapolation abilities to the near future, and frequent model updating or online learning would contribute to better predictive abilities for larger changes in the underlying process. 

As the above discussion reflects, data-driven VFMs are influenced by uncertainty. Both model (epistemic) uncertainty and measurement (aleatoric) uncertainty are present \cite{Hullermeier2021}. The first type originates from the model not being a perfect realization of the true process and there are uncertainties related to the model structure and parameters. The latter type is a cause of noisy data due to imprecision in measurements \cite{Gal2016}. Accounting for uncertainty is important to petroleum production engineers as they are often concerned with worst- and best-case scenarios. Further, information about the prediction uncertainty may aid the production engineers to decide whether the model predictions may be trusted. According to a recent survey \cite{Bikmukhametov2020a}, uncertainty estimation must be addressed by future research on VFM.

The motivation of this paper is to address uncertainty by introducing a probabilistic, data-driven VFM based on Bayesian neural networks. With this approach, epistemic uncertainty is modeled by considering the weights and biases of the neural network as random variables. Aleatoric uncertainty can be accommodated by a homoscedastic or heteroscedastic model of the measurement noise. This allows the modeler to separately specify priors related to the two uncertainty types. This can be beneficial when having knowledge of the measurement devices that produced the data modeled on.

Historically, the difficulty of performing Bayesian inference with neural networks has been a hurdle to practitioners. We thus provide a description of how to train the model using variational inference. Variational inference provides the means to perform efficient, approximate Bayesian inference and results in a posterior distribution over the model parameters \cite{Blei2017}. The method has shown promising results in terms of quantifying prediction uncertainty on other problems subject to small datasets and dataset shift \cite{Ovadia2019}. We also consider maximum a posteriori estimation, which serves as a non-probabilistic reference method. Although it computes a point estimate of the parameters, as opposed to a posterior distribution, it more closely resembles the maximum likelihood methods used in the majority of previous works on data-driven VFM. The reference method enables us to investigate if a probabilistic method, i.e. variational inference, may improve robustness over a non-probabilistic method. We test the proposed VFM by performing a large-scale empirical study on data from a diverse set of 60 petroleum wells. 

The paper is organized as follows. In Section \ref{sec:related-work} we briefly survey related works on data-driven VFM, with a focus on applications of neural networks. This section also gives some relevant background on probabilistic modeling. In Section \ref{sec:measurements-and-dataset} we describe how flow rates are measured and the dataset used in the case study. The probabilistic model for data-driven VFM is presented in Section \ref{sec:probabilistic-model} and in Section \ref{sec:methods} we discuss methods for Bayesian inference. The case study is presented in Section \ref{sec:case-study} and discussed in Section \ref{sec:discussion}. In Section \ref{sec:conclusion} we conclude and give our recommendations for future research on data-driven VFM based on our findings.

\section{Related work}
\label{sec:related-work}

\subsection{Traditional data-driven modeling}
\label{sec:intro-trad}
In literature, several data-driven methods have been proposed for VFM modeling, for instance, linear and nonlinear regression, principal component regression, random forest, support vector machines and the gradient boosting machine learning algorithm \cite{Zangl2014, Bello2014, Xu2011, Bikmukhametov2019}. One of the most popular and promising data-driven methods for VFM are neural networks (NN). In \cite{Zangl2014}, the oil flow rate from three wells was modeled using NNs, and an error as low as 0.15\% was reported. However, well-step tests were used to generate data with sufficient variety, and the time-span of the data covered only 30 hours. The three studies, \cite{Berneti2011, Ahmadi2013, Hasanvand2015}, investigated NNs for the oil flow rate from a reservoir using data samples from 31-50 wells. All used a neural network architecture with one hidden layer and 7 hidden neurons. In the two first, the imperialist competitive algorithm was used to find the NN weights. All of the three studies reported a very small mean squared error, of less than $0.05$. Yet, the data was limited to a time-span of 3 months and did not include measurements of the choke openings of the petroleum wells. This will strongly affect the future model performance when reservoir conditions change and the choke openings are adjusted.

A particularly noticeable series of studies on VFM and NN, using historical well measurements with a time-span of more than a year, are \cite{AlQutami2018, AlQutami2017b, AlQutami2017a, AlQutami2017c}. In \cite{AlQutami2017b}, the oil and gas flow rates were modeled using two individual feed-forward NN, with one hidden layer and 6 and 7 neurons respectively, and with early stopping to prevent overfitting. An error of 4.2\% and 2.3\% for the oil and gas flow rates were reported. In \cite{AlQutami2017c}, a radial basis function network was utilized to model the gas flow rate from four gas condensate wells, and the Orthogonal Least Squares algorithm was applied to find the optimal number of neurons ($\leq80$) in the hidden layer of the network. The study reported an error of 5.9\%. In \cite{AlQutami2017a, AlQutami2018}, ensemble neural networks were used to excel the learning from sparse data. In the first, the neural network architecture was limited to one hidden layer but the number of hidden neurons was randomly chosen in the range 3-15. Errors of 1.5\%, 6.5\%, and 4.7\% for gas, oil, and water flow rate predictions were achieved. The second paper considered 1-2 hidden layers with 1-25 neurons. Errors of 4.7\% and 2.4\% were obtained for liquid and gas flow rates respectively. 

\subsection{Probabilistic modeling}\label{sec:intro-prob}
A common approach in today's industry and literature is to study the sensitivity of the model to changes in parameter values, thus to a certain extent approaching epistemic uncertainty, e.g. \cite{Bieker2007, Fonseca2009, Zangl2014, Monteiro2017, Monteiro2020}. By approximating probability distributions for some of the model parameters from available process data and using sampling methods to propagate realizations of the parameters through the model, a predictive distribution of the output with respect to the uncertainty in the parameter may be analyzed.

Probabilistic modeling offers a more principled way to model uncertainty, e.g. by considering model parameters and measurement noise as random variables \cite{Ghahramani2015}. With Bayesian inference, a posterior distribution of the model output is found that takes into account both observed process data and prior beliefs of the model parameters \cite{Hastie2009}. The result is a predictive model that averages over all likely models that fit the data and a model that offers a natural parameter regularization scheme through the use of priors. This is in contrast to traditional data-driven modeling where the concern is often to find the maximum likelihood estimate \cite{Ghahramani2015}. Although probabilistic models and Bayesian inference are well-known in other fields of research, probabilistic VFMs are rare, yet existent \cite{Lorentzen2014, Lorentzen2016, Luo2014, Bassamzadeh2018}. 

The following series of studies, \cite{Lorentzen2014, Luo2014, Lorentzen2016}, constructed a mechanistic, probabilistic model of the flow rate in petroleum wellbores. A method for probabilistic, data-driven models is Bayesian neural networks (BNNs). BNNs are similar to traditional neural networks but with each parameter represented with a probability distribution \cite{Hastie2009, Polson2017}. Bayesian methods have shown to be efficient in finding high accuracy predictors in small data regimes and in the presence of measurement noise without overfitting to the data \cite{Snoek2012}. Further, Bayesian methods lend themselves to online model updating and could quickly improve the model's predictive ability when introduced to new operating regions. Yet, there are disadvantages with probabilistic modeling and Bayesian inference. Except in special cases, inferring the posterior probability distribution of the model consists of solving intractable integrals and inference is slow for large datasets \cite{Blei2017}. However, methods for approximation of the posterior distribution exist such as Markov Chain Monte Carlo (MCMC) sampling and variational inference (VI). Comparing these two approximation methods, VI has shown to scale better to large datasets and inference tends to be faster. Additionally, it simplifies posterior updating in the presence of new data. Nevertheless, the approximation with VI is in most cases bounded away from the true distribution, whereas MCMC methods will in principle converge towards the true distribution \cite{Blei2017}. A challenge for data-driven probabilistic models, such as Bayesian neural networks, is that the model parameters are generally non-physical, and setting the parameter priors is nontrivial. Despite neural networks being among the more popular data-driven methods for VFM modeling, to the extent of the authors' knowledge, there has been no attempt at using BNNs for VFM. There are, however, examples of BNNs being used for data-driven prediction in similar applications \cite{Liu2012,Humphrey2016}.

\section{Flow rate measurements and dataset}
\label{sec:measurements-and-dataset}
\begin{figure}[ht]
	\centering
	\includegraphics[width=1.0\linewidth]{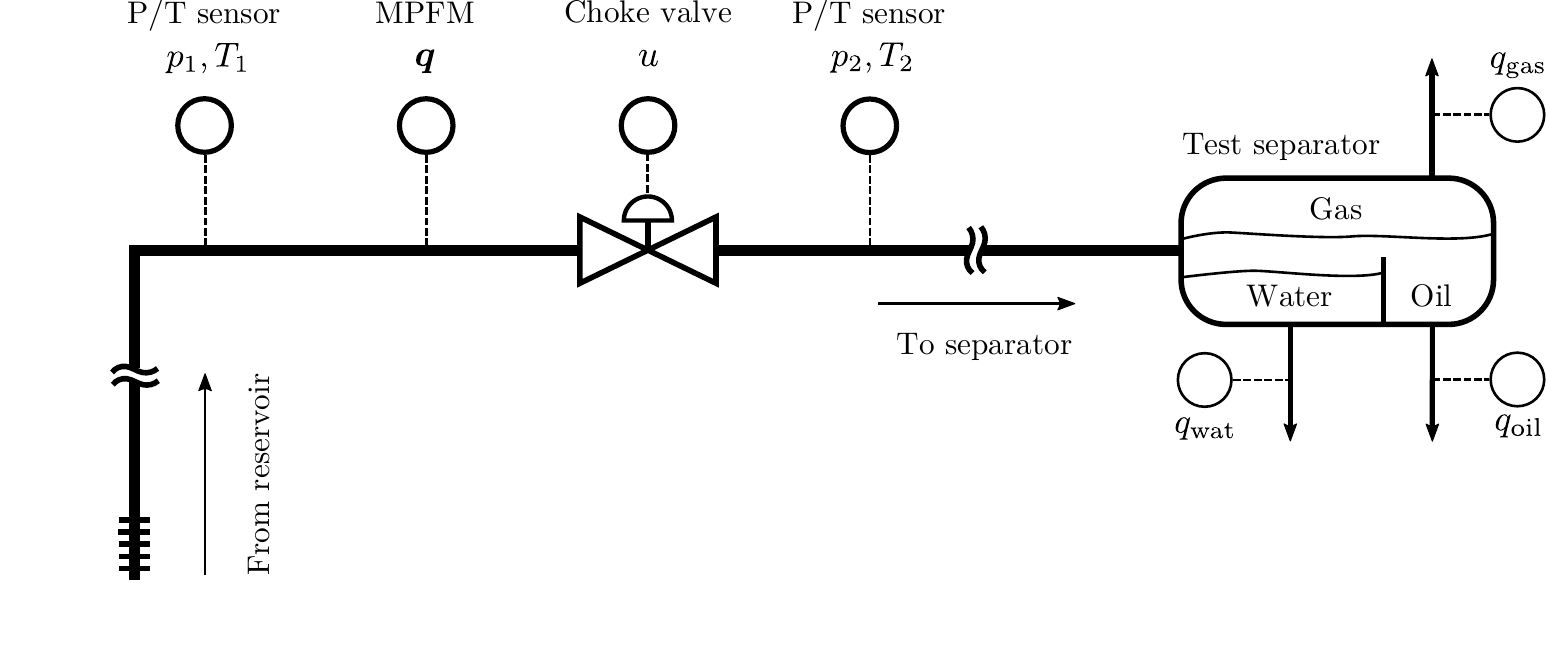}
	\caption{Sensor placement in a typical production well. A MPFM measures multiphase flow rates in the well. During well testing, single phase flow rates are measured with high accuracy after fluid separation.}
	\label{fig:well_sensors}
\end{figure}

A petroleum production well is illustrated in Figure \ref{fig:well_sensors}. Produced fluids flow from the reservoir, up to the wellhead, and through the choke valve. The choke valve opening ($u$) is operated to control the production from the well. The fluids thereafter enter the separator which separates the multiphase flow into the three single phases of oil, gas, and water $\bm{q} = (q_{\text{oil}}, q_{\text{gas}}, q_{\text{wat}})$. On well-instrumented wells, pressure ($p$) and temperature ($T$) is measured upstream and downstream the choke valve. 

The two main devices to measure multiphase flow rates in a well are the multiphase flow meter (MPFM) and test separator, both illustrated in Figure~\ref{fig:well_sensors}. MPFMs are complex devices based on several measurement principles and offer continuous measurements of the multiphase flow rate. Unfortunately, MPFMs have limited operation range, struggle with complex flow patterns, and are subject to drift over time \cite{Corneliussen2005}. Additionally, PVT (pressure-volume-temperature) data are used as part of the MPFM calculations and should be accurate and up-to-date for high accuracy MPFM measurements. On the other hand, well-testing is performed by routing the multiphase flow to a test separator whereby the separated flows are measured using single-phase measurement devices over a period of time (typically a few hours). Compared to the MPFM, well tests are performed infrequently, usually 1-2 times a month \cite{Monteiro2020}.

Normally, measurements of the multiphase flow rate obtained through well-testing have higher accuracy than the measurements from the MPFMs. This is due to the use of single-phase measurement devices in well-testing. According to \cite{Corneliussen2005, Marshall2015}, the uncertainty, in terms of mean absolute percentage error, of well tests, may potentially be as low as 2\% and 1\% for gas and oil respectively,  whereas MPFM uncertainty is often reported to be around 10\%. The error statistics are calculated with respect to reference measurements. For measurements of pressure, temperature, and choke openings, we assume that the sensors' accuracy is high, typically with an uncertainty of 1\% or less, and measurement error in these measurements are therefore neglected.   

The flow rates are often given as volumetric flow rates under standard conditions, e.g. as standard cubic meter per hour ($Sm^3/h$). Standard conditions make it easier to compare to reference measurements or measurements at other locations in the process as the volume of the fluid changes with pressure and temperature. Flow rates may be converted from actual conditions to standard conditions using PVT data \cite{Krejbjerg2019}. If the density of the fluid at standard conditions is known, the standard volumetric flow rate may be converted to mass flow rate, and the phasic mass fractions, $\bm{\eta} = (\eta_{\text{oil}}, \eta_{\text{gas}}, \eta_{\text{wat}})$, may be calculated. We assume steady-state production, frozen flow, and incompressible liquid such that the phasic volumetric flow rate and mass fractions are constant through the system, from the reservoir to the separator. 

\subsection{Dataset}
\label{sec:dataset}
The dataset used in this study consists of 66 367 data points from 60 wells producing from five oil and gas fields. The dataset was produced from raw measurement data using a data squashing technology \citep{Grimstad2016}. The squashing procedure averages raw measurement data in periods of steady-state operation to avoid short-scale instabilities. The resulting data points, which we refer to as measurements henceforth, are suitable for modeling of steady-state production rates.

\begin{figure}[ht]
\centering
\subfloat[Measurement frequency]{
\includegraphics[width=0.47\textwidth]{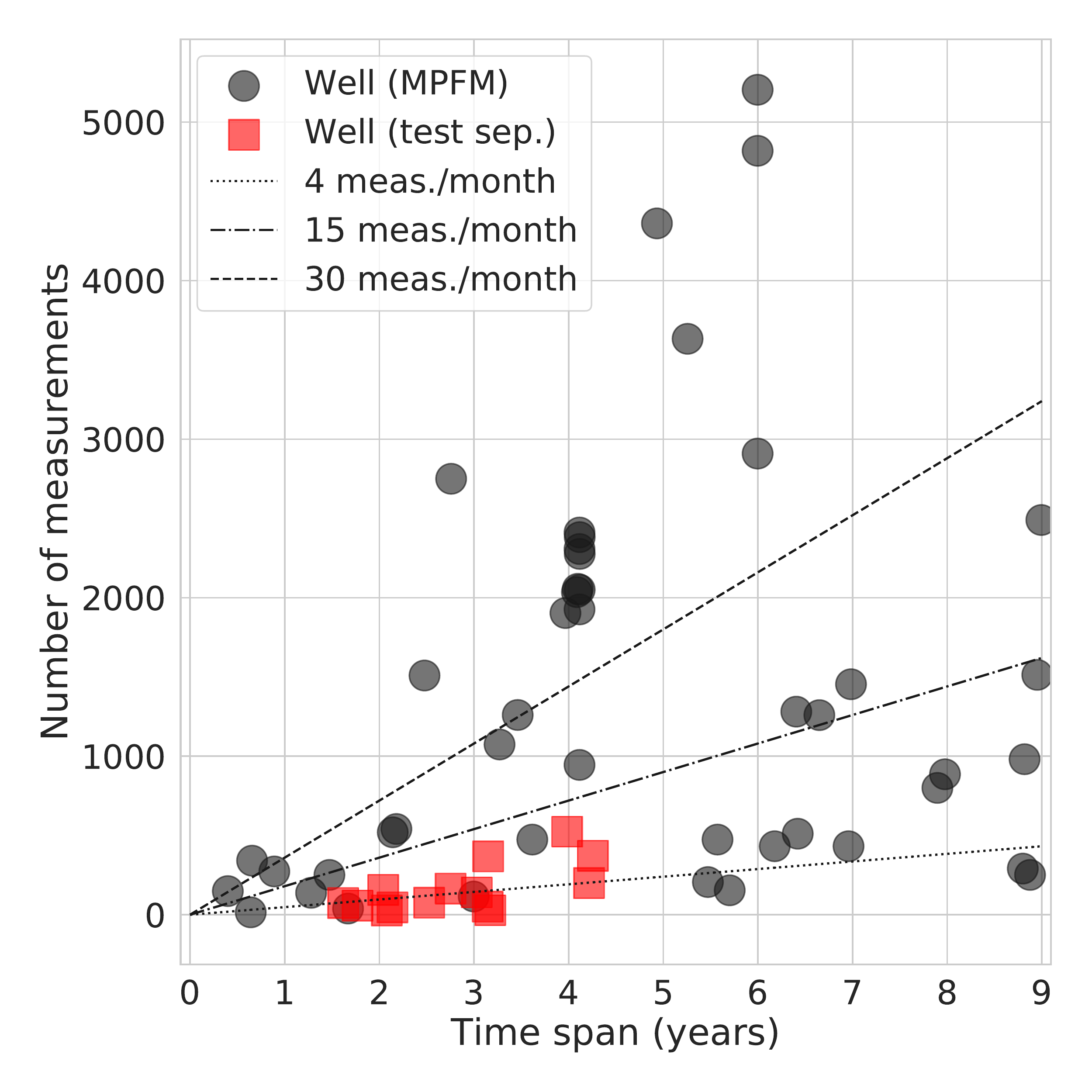}
\label{fig:well-experiments}
}
\hfill
\subfloat[Average mass fractions]{
\includegraphics[width=0.47\textwidth]{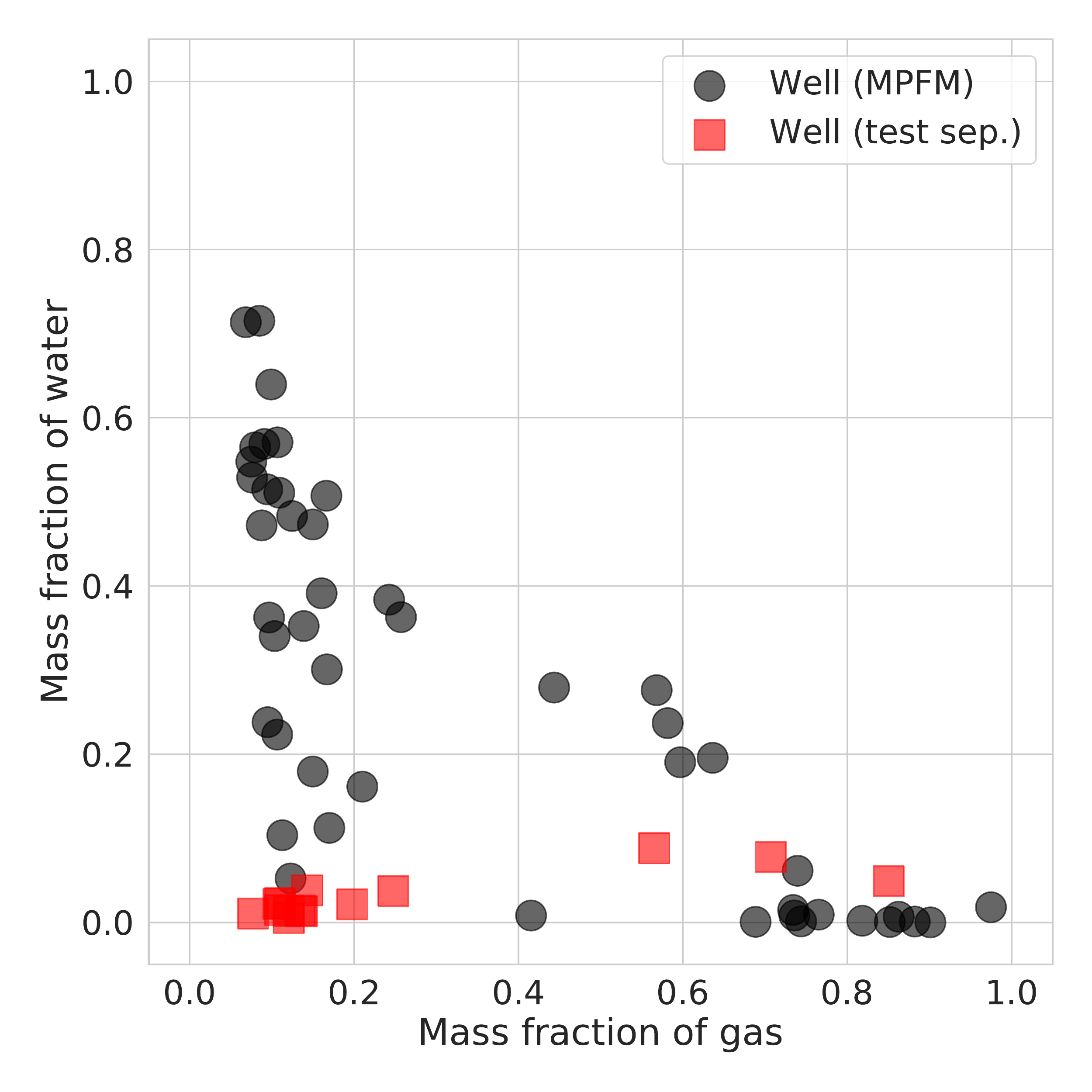}
\label{fig:fractions}
}
\caption{The number of measurements is plotted against the time span from the first to last measurement in (a). The average gas and water mass fraction is shown for all wells in (b).}
\label{fig:dataset}
\end{figure}

For each well we have a sequence of measurements in time. The time span from the first to last measurement is plotted for each well in Figure \ref{fig:well-experiments}. The figure shows that the measurement frequency varies from a handful to hundreds of measurements per year. There are 14 wells with test separator measurements, for which the average number of measurements is 163. The other 46 wells have MPFM measurements, and the average number of measurements is 1393. The 60 wells are quite different from each other in terms of produced fluids. Figure \ref{fig:fractions} illustrates the spread in mass fractions among the wells.

In the following, we model the multiphase flow through the production choke valve, a crucial component in any VFM. We consider ideal conditions, in the sense that all measurements required by a reasonable choke model are available \cite{Hotvedt2020}. For each well, we collect the corresponding measurements in a dataset $\dataset = \left\{(\bm{x}_i, y_i)\right\}_{i=1}^{N}$. We will only consider one well at the time and simply refer to the dataset as $\dataset$. The target variable is the total volumetric flow rate, $y_i = q_{\text{oil}, i} + q_{\text{gas}, i} + q_{\text{wat}, i} \in \reals$, measured \textit{either} by a test separator \textit{or} a MPFM. The explanatory variables, 
\begin{equation*}
\bm{x}_i = (u_i, p_{1,i}, p_{2,i}, T_{1,i}, T_{2,i}, \eta_{\text{oil},i}, \eta_{\text{gas},i}) \in \reals^7,
\end{equation*}
are the measured choke opening, the pressures and temperatures upstream and downstream the choke valve, and the mass fractions of oil and gas. No experimental set-up was used to affect the data variety; for example, we did not consider step well tests as in \cite{Zangl2014}.

\section{Probabilistic flow model}
\label{sec:probabilistic-model}
Consider the following probabilistic model for the total multiphase flow rate:
\begin{equation}
\begin{aligned}
&\left.
\begin{aligned}
y_i &= z_i + \epsilon_i \\
z_i &= f(\bm{x}_i, \bm{\phi}) \\
s_i &= g(z_i, \bm{\psi}) \\
\epsilon_i &\sim \normaldist{}(0, s_i^2) \\
\end{aligned}
\right\rbrace i=1,\ldots,N, \\
&~
\begin{aligned}
\bm{\phi} &\sim p(\bm{\phi}) = \prod\limits_{i=1}^{K_{\phi}} \normaldist{}(\phi_i \condbar a_i, b_i^2), \\
\bm{\psi} &\sim p(\bm{\psi}) = \prod\limits_{i=1}^{K_{\psi}} \normaldist{}(\psi_i \condbar c_i, d_i^2), \\
\end{aligned}
\end{aligned}
\label{eq:flow-model}
\end{equation}
where $y_i$ is a measurement of the multiphase flow rate $z_i$ subject to additive measurement noise $\epsilon_i$. The nonlinear dependence of $z_i$ on $\bm{x}_i$ is approximated by a Bayesian neural network $f(\bm{x}_i, \bm{\phi})$ with weights and biases represented by latent (random) variables $\bm{\phi}$. The neural network is composed of $L$ functions, $f = f^{(L)} \circ \cdots \circ f^{(1)}$, where $f^{(1)}$ to $f^{(L-1)}$ are called the hidden layers of $f$, and $f^{(L)}$ is the output layer \cite{Goodfellow2016}. A commonly used form of a hidden layer $l$ is $f^{(l)}(\bm{x}) = \text{ReLU}(W^{(l)} \bm{x} + \bm{b}^{(l)})$, where the rectified linear unit (ReLU) operator is given as $\text{ReLU}(\bm{z})_i = \max \{z_i, 0\}$, $W^{(l)}$ is a weight matrix, and $\bm{b}^{(l)}$ is a vector of biases. For regression tasks the output layer is usually taken to be an affine mapping, $f^{(L)}(\bm{x}) = W^{(L)} \bm{x} + \bm{b}^{(L)}$. The layer weights and biases are collected in $\bm{\phi} = \{(W^{(l)}, \bm{b}^{(l)})\}_{l=1}^{L}$ to enable the compact notation $f(\bm{x}_i, \bm{\phi})$. With a slight abuse of this notation, an element $\phi_i$ of $\bm{\phi}$ represents a scalar weight or bias for $i \in \{1, \ldots, K_{\phi}\}$, where $K_{\phi}$ is the total number of weights and biases in the neural network. The distinguishing feature of a Bayesian neural network is that the weights and biases, $\bm{\phi}$, are modeled as random variables with a prior distribution $p(\bm{\phi})$.

\begin{figure}[ht]
\centering
\includegraphics[width=0.4\linewidth]{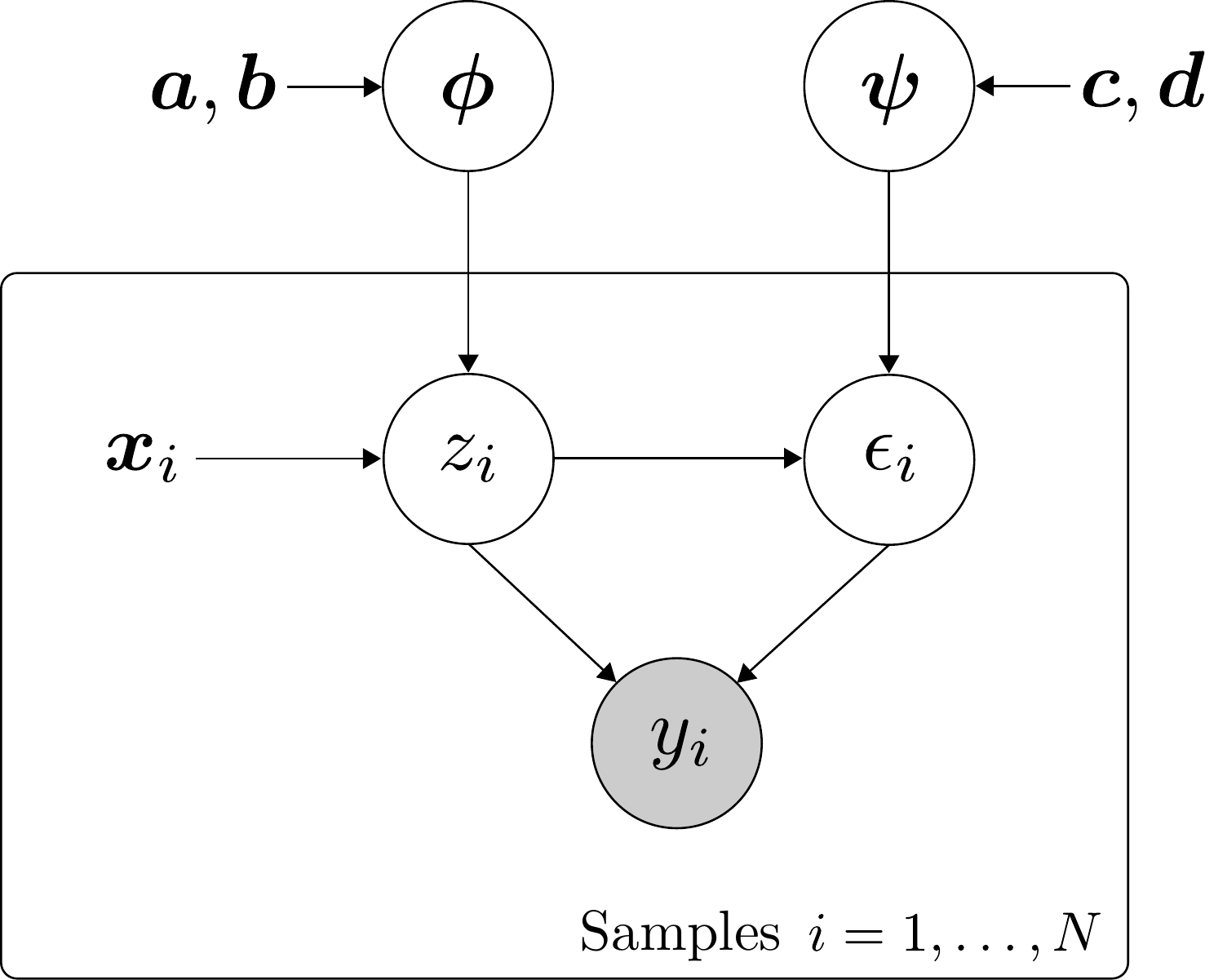}
\caption{A probabilistic graphical model for flow rates. Random variables are inscribed by a circle. A gray-filled circle means that the random variable is observed. The dependence $z_i \to \epsilon_i$ indicates that the noise is heteroscedastic, while the dependence $\bm{\psi} \to \epsilon_i$ indicates that the noise model is learned from data.}
\label{fig:pgm-flow-rate}
\end{figure}

We assume the noise to be normally distributed with standard deviation $g(z_i, \bm{\psi}) > 0$, and we consider different functions $g$ of $z_i$ and latent variables $\bm{\psi}$. We discuss the priors on the latent variables, $p(\bm{\phi})$ and $p(\bm{\psi})$, in the subsequent sections. The probabilistic model is illustrated graphically in Figure \ref{fig:pgm-flow-rate}.

Given $\bm{\phi}$, $\bm{\psi}$ and explanatory variables $\bm{x}$, the conditional flow rate $z = f(\bm{x}, \bm{\phi})$ and a measurement $y$ is generated as
\begin{equation}
y \condbar z, \bm{\psi}  \sim \mathcal{N}(y \condbar z, g(z, \bm{\psi})^2).
\end{equation}
The flow rate measurement $y$ is subject to epistemic (model) uncertainty in $f(\bm{x}, \bm{\phi})$ and aleatoric (measurement) uncertainty via $g(z, \bm{\psi})$. We differ between homoscedastic and heteroscedastic measurement noise. Heteroscedasticity is when the structure of the noise in a signal is dependent on the structure of the signal itself and is more difficult to capture \cite{Woodward1998}. Homoscedasticity is the lack of heteroscedasticity.

The flow model in \eqref{eq:flow-model} is a quite generic regression model, but it restricts the modeling of the measurement noise. The model allows the noise to be heteroscedastic, with the noise level being a function of the flow rate $z$, or homoscedastic for which the noise level is fixed. In the latter case, $g(z, \bm{\psi}) = \sigma_n$, where $\sigma_n$ is a fixed noise level. If the noise level is unknown, it can be learned with the following homoscedastic noise model:
\begin{equation}
\begin{aligned}
g(z_i, \bm{\psi}) &= \exp(\psi_1), \\
\psi_1 &\sim \normaldist{}(c_1, d_1^2),
\end{aligned}
\label{eq:homo-noise-model}
\end{equation}
where $\psi_1$ is a normally distributed latent variable and the noise level is log-normal. The exponential ensures that $g(z_i, \bm{\psi}) > 0$.

The homoscedastic noise model in \eqref{eq:homo-noise-model} may be unrealistic for flow meters with a heteroscedastic noise profile. As described earlier, the uncertainty of the flow rate measurement is often given in relative terms. To model this property of the data, we augment \eqref{eq:homo-noise-model} with a multiplicative term to get the following heteroscedastic noise model:
\begin{equation}
\begin{aligned}
g(z_i, \bm{\psi}) &= \exp(\psi_2) \cdot |z_i| + \exp(\psi_1), \\ 
\psi_1 &\sim \normaldist{}(c_1, d_1^2), \\
\psi_2 &\sim \normaldist{}(c_2, d_2^2),
\end{aligned}
\label{eq:hetero-noise-model}
\end{equation}
where $\psi_1$ and $\psi_2$ are normally distributed latent variables\footnote{We assume that we have one flow rate instrument for each well. Yet, several instruments may be handled by having separate noise models for each instrument.}. Both $\exp(\psi_1)$ and $\exp(\psi_2)$ are log-normal, and are hence strictly positive. It follows from $|z| \geq 0$ that the noise standard deviation $g(z, \bm{\psi}) > 0$. 

\subsection{Prior for the noise model, $p(\bm{\psi})$}
\label{sec:prior-noise-model}
The prior for the noise model is assumed to be a factorized normal 
\begin{equation}
p(\bm{\psi}) = \prod\limits_{i=1}^{K_{\psi}} \normaldist{}(\psi_i \condbar c_i, d_i^2),
\end{equation}
where $K_{\psi} = 1$ for the homoscedastic noise model in \eqref{eq:homo-noise-model} and $K_{\psi} = 2$ for the heteroscedastic noise model in \eqref{eq:hetero-noise-model}.

The accuracy of an instrument measuring flow rate is commonly given as a mean absolute percentage error (MAPE) to a reference measurement. More precisely, the expected measurement error is specified as
\begin{equation}
    \expectationp[y \condbar z]{\frac{|y - z|}{|z|}} = E_r,
    \label{eq:measurement-error}
\end{equation}
where $y$ is the measurement, $z > 0$ is the reference measurement, and $E_r$ is the MAPE, e.g. $E_r = 0.1$ for a MAPE of 10\%. We wish to translate such statements to a prior $p(\bm{\psi})$. 

Assuming a perfect reference measurement $z$, normal noise $\epsilon$, and an additive noise model $y = z + \epsilon$, we obtain from \eqref{eq:measurement-error} a noise standard deviation $g(z) = \sqrt{\pi/2} E_r |z|$. We recognize this as the first term in the heteroscedastic noise model \eqref{eq:hetero-noise-model}. We derive prior parameters of $\psi_2 \sim \normaldist{}(c_2, d_2^2)$ that correspond to a log-normal distribution $\exp{}(\psi_2)$ with mean $\sqrt{\pi/2} E_r$ by solving:
\begin{equation}
    c_2 = \log(\sqrt{\pi/2} E_r) - d_2^2 / 2, 
\end{equation}
where we can adjust the variance $d_2^2$ to express our uncertainty in the value of $E_r$.

The specification of a relative measurement error $E_r$ cannot be translated directly to a fixed noise level, as required by the homoscedastic noise model in \eqref{eq:homo-noise-model}. However, we can obtain a reasonable approximation by using the above procedure. If we set $z = \bar{z}$, where $\bar{z}$ is the mean production of a well, we can calculate prior parameters for $\psi_1$ as follows:
\begin{equation}
    c_1 = \log(\sqrt{\pi/2} E_r \bar{z}) - d_1^2 / 2.
\end{equation}
We express our uncertainty about the noise level by adjusting the variance $d_1^2$.


\subsection{Prior for the neural network weights, $p(\bm{\phi})$}
\label{sec:prior-neural-network}
We encode our initial belief of the parameters $\bm{\phi}$ with a fully factorized normal prior
\begin{equation}
p(\bm{\phi}) = \prod\limits_{i=1}^{K_{\phi}} \normaldist{}(\phi_i \condbar a_{i}, b_{i}^2),
\end{equation}
where $K_{\phi}$ is the number of weights and biases in the neural networks $f$. We assume a zero mean for the weights and biases, that is $a_i = 0$, as is common practice for neural networks. One interpretation of the prior standard deviations is that they encode the (believed) frequencies of the underlying function, with low values of $\bm{b}$ inducing slow-varying (low frequency) functions, and high values inducing fast-varying (high frequency) functions \cite{Gal2016}. While this interpretation can give us some intuition about the effect of the prior, it is not sufficiently developed to guide the specification of a reasonable prior. We refrain from learning the prior from the data (as with empirical Bayes) and therefore treat $\bm{b}$ as hyperparameters to be prespecified. 

For deep neural networks it is common practice to randomly sample the initial weights so that the output has a variance of one for a standard normal distributed input \cite{Glorot2010,He2015}. For example, He-initialization \cite{He2015} is often used for neural networks with ReLU activation functions. With He-initialization, the weights of layer $l$ are drawn from the distribution $\normaldist(0, \sigma_{l}^{2})$ with $\sigma_l = \sqrt{2/n_{l}}$, where $n_{l}$ is the number of layer inputs. The weights in the first hidden layer are initialized with $\sigma_l = \sqrt{1/n_{l}}$ since no ReLU activation is applied to the network's input. With layer biases set to zero, this initialization scheme yields a unit variance for the output.

The objective of weight initialization is similar to that of prior specification; a goal in both settings is to find a good initial model. In this work, we use the standard deviations $b_i = \sigma_l$ as a starting point for the prior specification (for weight $i$ in layer $l$ of a ReLU network). We call this the He-prior. The resulting standard deviations can then be increased (or decreased) if one believes that the underlying function amplifies (or diminishes) the input signal. 

\begin{figure}[H]
\centering
\subfloat[Two hidden layers, each with 50 nodes]{
\includegraphics[width=0.47\textwidth]{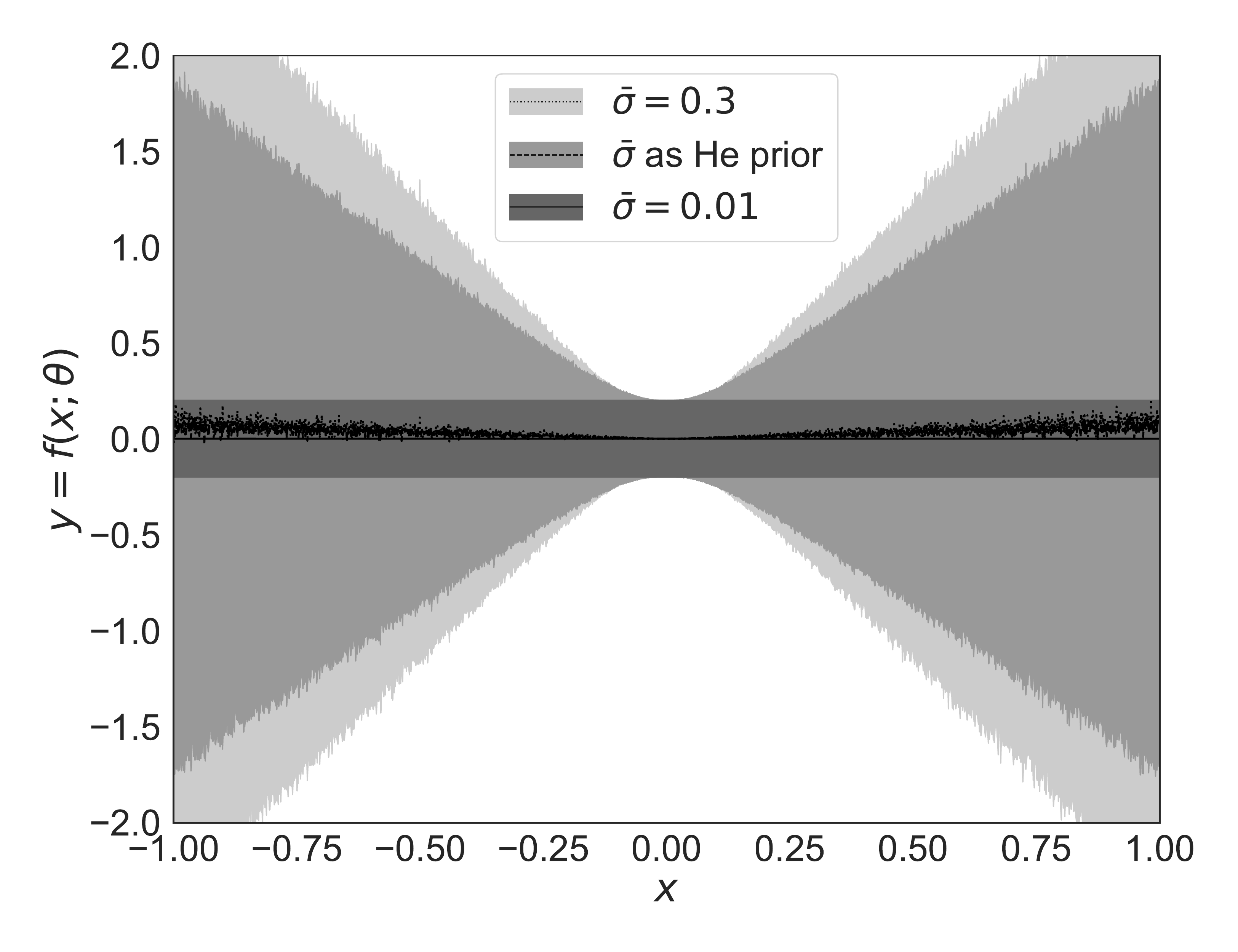}
\label{fig:priors-shallow}
}
\hfill
\subfloat[Five hidden layers, each with 50 nodes]{
\includegraphics[width=0.47\textwidth]{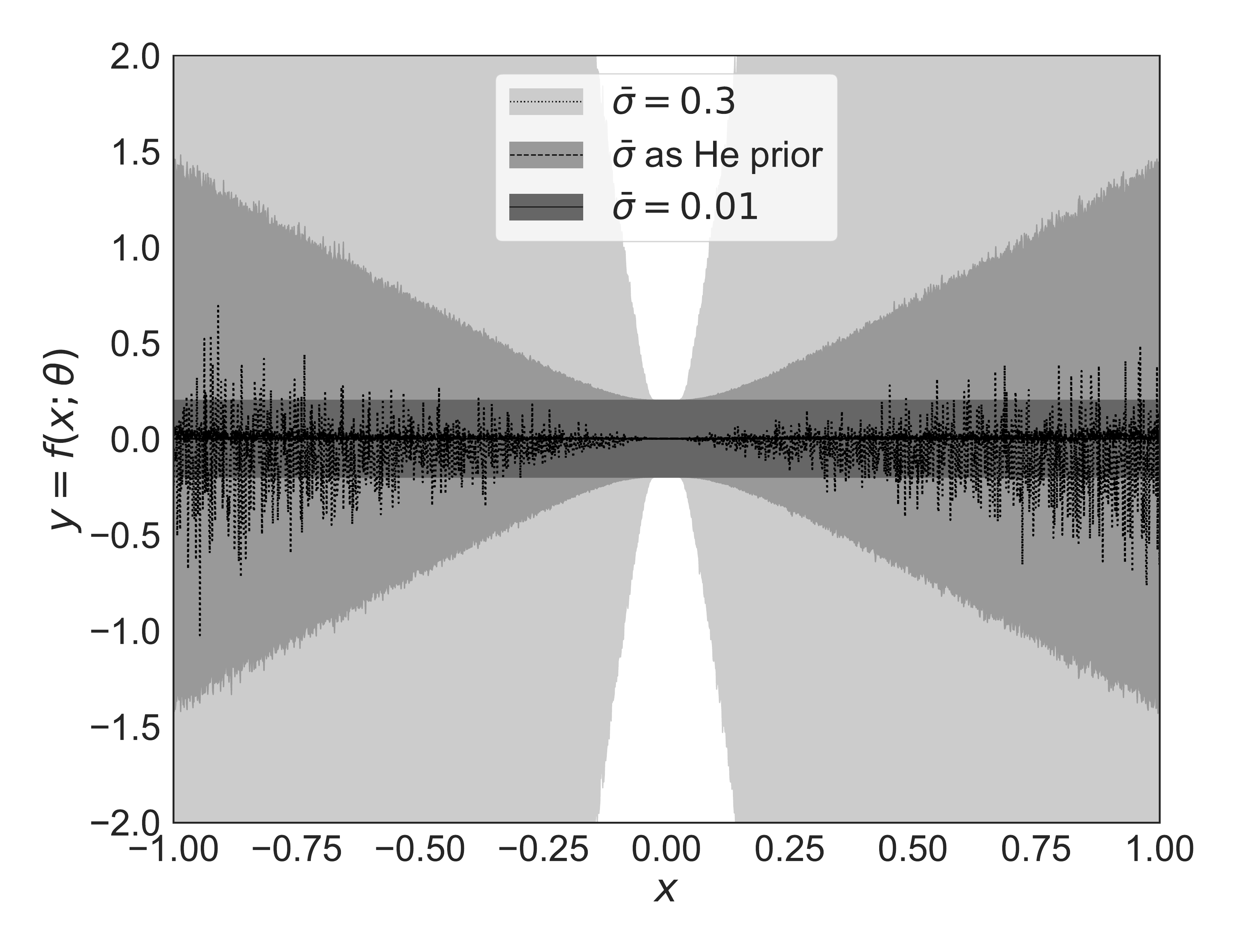}
\label{fig:priors-deep}
}
\caption{Prediction uncertainty (two sigma) for different priors $b_i = \bar{\sigma}$ on a neural network's weights. Two networks are trained on a dataset $\dataset = \{(0, y_i)\}_{i=1}^{100}$, where $y_i \sim \normaldist(0, \sigma_n^2)$ and the noise level $\sigma_n = 0.1$ is known. The figure shows that the epistemic (model) uncertainty is explained away for $x=0$ and increasing with the distance to $x=0$. Away from the data, the increase in epistemic uncertainty depends on the prior variance and network depth.}
\label{fig:priors}
\end{figure}

Figure \ref{fig:priors} shows the effect of $\bm{b}$ on the predictive uncertainty of a Bayesian neural network. With a common prior standard deviation (same for all weights), the output variance is sensitive to the network size (depth and width). This sensitivity complicates the prior specification, as illustrated for different network depths in the figure. The He-prior retains a unit output variance for different network sizes.

\subsection{A fully factorized normal prior on the latent variables}
The prior of model \eqref{eq:flow-model} is a fully factorized normal distribution, $p(\bm{\phi}) p(\bm{\psi})$. To simplify the notation in the rest of this paper we collect the latent variables in $\bm{\theta} = (\bm{\phi}, \bm{\psi}) \in \reals^K$, where $K = K_{\phi} + K_{\psi}$. This allows us to state the prior on $\bm{\theta}$ as $p(\bm{\theta}) = p(\bm{\phi}) p(\bm{\psi})$, where
\begin{equation}
p(\bm{\theta}) = \prod\limits_{i=1}^{K} \normaldist{}(\theta_i \condbar \bar{\mu}_i, \bar{\sigma}_{i}^{2}),
\label{eq:prior}
\end{equation}
with means $\bar{\bm{\mu}} = (\bar{\mu}_1, \ldots, \bar{\mu}_K) = (a_1, \ldots, a_{K_{\phi}}, c_{1}, \ldots, c_{K_{\psi}}) \in \reals^K$ and standard deviations $\bar{\bm{\sigma}} = (\bar{\sigma}_1, \ldots, \bar{\sigma}_K) = (b_1, \ldots, b_{K_{\phi}}, d_{1}, \ldots, d_{K_{\psi}}) \in \reals^K$. The total number of model parameters ($\bar{\bm{\mu}}$ and $\bar{\bm{\sigma}}$) is $2K$.

\section{Methods}
\label{sec:methods}

We wish to infer the latent variables $\bm{\theta}$ of the flow rate model in \eqref{eq:flow-model} from observed data. With Bayesian inference, the initial belief of $\bm{\theta}$, captured by the prior distribution $p(\bm{\theta})$ in \eqref{eq:prior}, is updated to a posterior distribution $p(\bm{\theta} \condbar \dataset)$ after observing data $\dataset$. The update is performed according to Bayes' rule:
\begin{equation}
\label{eq:posterior}
p(\bm{\theta} \condbar \dataset) = \frac{p(\dataset \condbar \bm{\theta}) p(\bm{\theta})}{p(\dataset)},
\end{equation}
where $p(\dataset)$ is the evidence and the likelihood is given by
\begin{equation}
p(\dataset \condbar \bm{\theta}) = \prod\limits_{i=1}^{N} p(y_i \condbar \bm{x}_i, \bm{\theta}).
\label{eq:likelihood}
\end{equation}
The log-likelihood of the model in \eqref{eq:flow-model} is shown in \ref{app:log-likelihood}.

From the posterior distribution, we can form the predictive posterior distribution
\begin{equation}
p(y^{\smallplus} \condbar \bm{x}^{\smallplus}, \dataset) = \int p(y^{\smallplus} \condbar \bm{x}^{\smallplus}, \bm{\theta}) p(\bm{\theta} \condbar \dataset) d\bm{\theta}
\label{eq:predictive-dist}
\end{equation}
to make a prediction $y^{\smallplus}$ for a new data point $\bm{x}^{\smallplus}$. 

The posterior in \eqref{eq:posterior} involves intractable integrals that prevents a direct application of Bayes' rule \citep{Blei2017}. In the following sections, we review two methods that circumvent this issue, namely maximum a posteriori (MAP) estimation and variational inference. With MAP estimation inference is simplified by considering only the mode of $p(\bm{\theta} \condbar \dataset)$, and with variational inference the posterior distribution is approximated. In the latter case, we can form an approximated predictive posterior distribution by replacing the posterior in \eqref{eq:predictive-dist} with its approximation. Statistics of this distribution, such as the mean and variance, can be estimated using Monte-Carlo sampling \cite{Gal2016}.

\subsection{MAP estimation}
\label{sec:map-estimation}
With maximum a posteriori (MAP) estimation we attempt to compute:
\begin{equation}
    \hat{\bm{\theta}}_{\text{MAP}} = \arg \max_{\bm{\theta}} ~ p(\bm{\theta} \condbar \mathcal{D}),
\end{equation}
where $\hat{\bm{\theta}}_{\text{MAP}}$ is the mode of the posterior distribution in \eqref{eq:posterior}. For the model in \eqref{eq:flow-model} with a fixed and constant noise variance $\sigma_n^2$ and $\bar{\sigma}_i^2$ is the (prior) variance of $\theta_i$, we have that 
\begin{equation}
\begin{aligned}
    \hat{\bm{\theta}}_{\text{MAP}} &= \arg \max_{\bm{\theta}} ~\log p( \mathcal{D} \condbar \bm{\theta}) + \log p(\bm{\theta}) \\
    &= \arg \min_{\bm{\theta}} ~ \frac{1}{2 \sigma_n^2} \sum_{i=1}^N \left(y_i - f(\bm{x}_i, \bm{\theta})\right)^2 + \sum_{i=1}^K \frac{1}{2\bar{\sigma}_{i}^2}\theta_i^2,
\end{aligned}
\label{eq:map-estimation-1}
\end{equation}
From \eqref{eq:map-estimation-1}, we see that MAP estimation is equivalent to maximum likelihood estimation with $L^2$-regularization \citep{Hastie2009}. 

While MAP estimation allows us to incorporate prior information about the model, it provides only a point estimate $\hat{\bm{\theta}}_{\text{MAP}}$ and will not capture the epistemic uncertainty of the model. To obtain a posterior distribution of $\bm{\theta}$ we consider the method of variational inference.

\subsection{Variational inference}
\label{sec:variational-inference}
With variational inference, the posterior in \eqref{eq:posterior} is approximated by solving an optimization problem, cf. \citep{Blei2017}. Consider a variational posterior density $q(\bm{\theta} \condbar \bm{\lambda})$, parameterized by a real vector $\bm{\lambda}$. The objective of the optimization is to find a density $q^\star = q(\bm{\theta} \condbar \bm{\lambda}^\star)$ that minimizes the Kullback-Leibler (KL) divergence to the exact posterior, i.e.
\begin{equation}
\label{eq:vi-direct}
\bm{\lambda}^\star = \underset{\bm{\lambda}}{\arg\min} ~\kullb{q(\bm{\theta} \condbar \bm{\lambda})}{p(\bm{\theta} \condbar \dataset)}.
\end{equation}

A direct approach to solve \eqref{eq:vi-direct} is not practical since it includes the intractable posterior. In practice, the KL divergence is instead minimized indirectly by maximizing the evidence lower bound (ELBO):
\begin{align}
\elbo{\bm{\lambda}} &:= \log p(\dataset) - \kullb{q(\bm{\theta} \condbar \bm{\lambda})}{p(\bm{\theta} \condbar \dataset)} \\
&~= \expectationp[q]{\log p(\dataset | \bm{\theta})} - \kullb{q(\bm{\theta} \condbar \bm{\lambda})}{p(\bm{\theta})},
\label{eq:elbo}
\end{align}
where the expectation $\expectationp[q]{\cdot}$ is taken with respect to $q(\bm{\theta} \condbar \bm{\lambda})$. From the ELBO loss in \eqref{eq:elbo}, we see that an optimal variational distribution maximizes the expected log-likelihood on the dataset, while obtaining similarity to the prior via the regularizing term $\kullb{q(\bm{\theta} \condbar \bm{\lambda})}{p(\bm{\theta})}$. 

\subsubsection{Stochastic gradient variational Bayes}
\label{sec:sgvb}
Stochastic gradient variational Bayes (SGVB) or Bayes by backprop is an efficient method for gradient-based optimization of the ELBO loss in \eqref{eq:elbo}, cf. \citep{Kingma2014, Blundell2015}. 

Suppose that the variational posterior $q(\bm{\theta} \condbar \bm{\lambda})$ is a mean-field (diagonal) normal distribution with mean $\bm{\mu}$ and standard deviation $\bm{\sigma}$. Let the variational parameters be $\bm{\lambda} = (\bm{\mu}, \bm{\rho})$ and compute $\bm{\sigma} = \log(1 + \exp(\bm{\rho}))$, where we use an elementwise softplus mapping to ensure that $\sigma_i > 0$. 

The basic idea of SGVB is to reparameterize the latent variables to $\bm{\theta} = h(\bm{\zeta}, \bm{\lambda}) = \bm{\mu} + \log(1 + \exp(\bm{\rho})) \circ \bm{\zeta}$, where $\circ$ denotes pointwise multiplication and $\bm{\zeta} \sim \mathcal{N}(0, I)$. With this formulation, the stochasticity of $\bm{\theta}$ is described by a standard normal noise $\bm{\zeta}$ which is shifted by $\bm{\mu}$ and scaled by $\bm{\sigma}$. The reparameterization allows us to compute the gradient of the ELBO \eqref{eq:elbo} as follows:
\begin{align}
\nabla_{\bm{\lambda}} \elbo{\bm{\lambda}} &= \nabla_{\bm{\lambda}} \expectationp[q]{\log p(\dataset | \bm{\theta})} - \nabla_{\bm{\lambda}} \kullb{q(\bm{\theta} \condbar \bm{\lambda})}{p(\bm{\theta})} \notag \\
&= \expectationp[\zeta]{\nabla_{\bm{\theta}} \log p(\dataset \condbar \bm{\theta}) \nabla_{\bm{\lambda}} h(\bm{\zeta}, \bm{\lambda})} - \nabla_{\bm{\lambda}} \kullb{q(\bm{\theta} \condbar \bm{\lambda})}{p(\bm{\theta})}
\label{eq:elbo-grad}
\end{align}

The expectation in \eqref{eq:elbo-grad} can be approximated by Monte-Carlo sampling the noise: $\bm{\zeta}_i \sim \normaldist{}(0, I)$ for $i = 1, \ldots, M$. If we also approximate the likelihood by considering a mini-batch $\minibatch \subset \dataset$ of size $B \leq N$, we obtain the unbiased SGVB estimator of the ELBO gradient:
\begin{equation}
\begin{aligned}
\nabla_{\bm{\lambda}} \elbo{\bm{\lambda}} \simeq \nabla_{\bm{\lambda}} \hat{\mathcal{L}}(\bm{\lambda}) &:= \frac{N}{B} \frac{1}{M} \sum\limits_{i=1}^{M} \nabla_{\bm{\theta}} \log p(\minibatch \condbar \bm{\theta}) \nabla_{\bm{\lambda}} h(\bm{\zeta}_i, \bm{\lambda}) \\ &\qquad - \nabla_{\bm{\lambda}} \kullb{q(\bm{\theta} \condbar \bm{\lambda})}{p(\bm{\theta})}.
\label{eq:elbo-grad-estimator}
\end{aligned}
\end{equation}

An advantage with the SGVB estimator in \eqref{eq:elbo-grad-estimator} is that we can utilize the gradient of the model $\nabla_{\bm{\theta}} \log p(\minibatch | \bm{\theta})$ as computed by back-propagation. When both the variational posterior and prior are mean-field normals, as is the case for our model, $\kullb{q(\bm{\theta} \condbar \bm{\lambda})}{p(\bm{\theta})}$ can be computed analytically as shown in \ref{app:KL-term}.

In Algorithm \ref{alg:sgvb} we summarize the basic SGVB algorithm for mean-field normals and Monte-Carlo sample size of $M=1$. We finally note that for variables representing weights of a neural network, we implement the local reparameterization trick in \cite{Kingma2015} to reduce gradient variance and save computations (not shown in Algorithm \ref{alg:sgvb}).

\begin{algorithm}[ht]
\caption{Basic implementation of SGVB for mean-field normals ($M=1$)}
\label{alg:sgvb}
\begin{algorithmic}[1]
\Require data $\dataset$, model $p(\dataset, \bm{\theta}) = p(\dataset \condbar \bm{\theta}) p(\bm{\theta})$, parameters $\bm{\lambda} = (\bm{\mu}, \bm{\rho})$, learning rate $\alpha$.
\Repeat
\State Sample mini-batch $\mathcal{B}$ from $\dataset$
\State Sample $\bm{\zeta} \sim \normaldist{}(0, I)$
\State $\bm{\theta} \gets \bm{\mu} + \log(1 + \exp(\bm{\rho})) \circ \bm{\zeta}$
\State Compute $\nabla_{\bm{\lambda}} \hat{\mathcal{L}}(\bm{\lambda})$ using \eqref{eq:elbo-grad-estimator}
\State $\bm{\lambda} \gets \bm{\lambda} + \alpha \nabla_{\bm{\lambda}} \hat{\mathcal{L}}(\bm{\lambda})$
\Until{no improvement in ELBO} \\
\Return $\bm{\lambda}$
\end{algorithmic}
\end{algorithm}

\section{Case study}
\label{sec:case-study}
The goal of the case study was to investigate the predictive performance and generalization ability of the proposed VFM. The study was designed to test the predictive performance on historical data and on future data, which reflect the two main applications of a VFM. If the models generalize well, a similar performance across all wells for each model type should be expected on both historical and future data. To cast light on the data challenges in Section \ref{sec:introduction}, the results differentiate between wells with test separator and MPFM measurements, which have different measurement accuracy and frequency. The prediction uncertainty of the models was also analyzed and the effect of training set size on prediction performance was investigated. 

The probabilistic flow rate models in Section \ref{sec:probabilistic-model} were developed using the dataset described in Section \ref{sec:dataset}. The conditional mean flow rate, $f(\bm{x}, \bm{\phi})$, was modeled using a feed-forward neural network. Three different noise models were considered: a homoscedastic model with fixed noise standard deviation $g(z, \bm{\psi}) = \sigma_n = \text{const.}$, a homoscedastic model with learned noise standard deviation \eqref{eq:homo-noise-model}, and a heteroscedastic model with learned noise standard deviation \eqref{eq:hetero-noise-model}. For each of the three model types and the 60 wells in the dataset, the neural network was trained using the SGVB method in Section \ref{sec:sgvb}. These models will be referred to by the label VI-NN. For comparison, a neural network for each of the 60 wells was trained using the MAP estimation method in Section \ref{sec:map-estimation}. For these models we considered the measurement noise to be homoscedastic with a fixed noise standard deviation ($\sigma_n$). We label these models as MAP-NN. The He-prior was used for the hidden layers to initialize and regularize the parameters, see Section \ref{sec:prior-neural-network}. For the noise models, we set the priors as described in Section \ref{sec:prior-noise-model}, differentiating between wells with MPFM and test separator measurements. 

\begin{figure}[H]
\centering
\subfloat[Architecture of BNN]{
\includegraphics[width=0.7\textwidth]{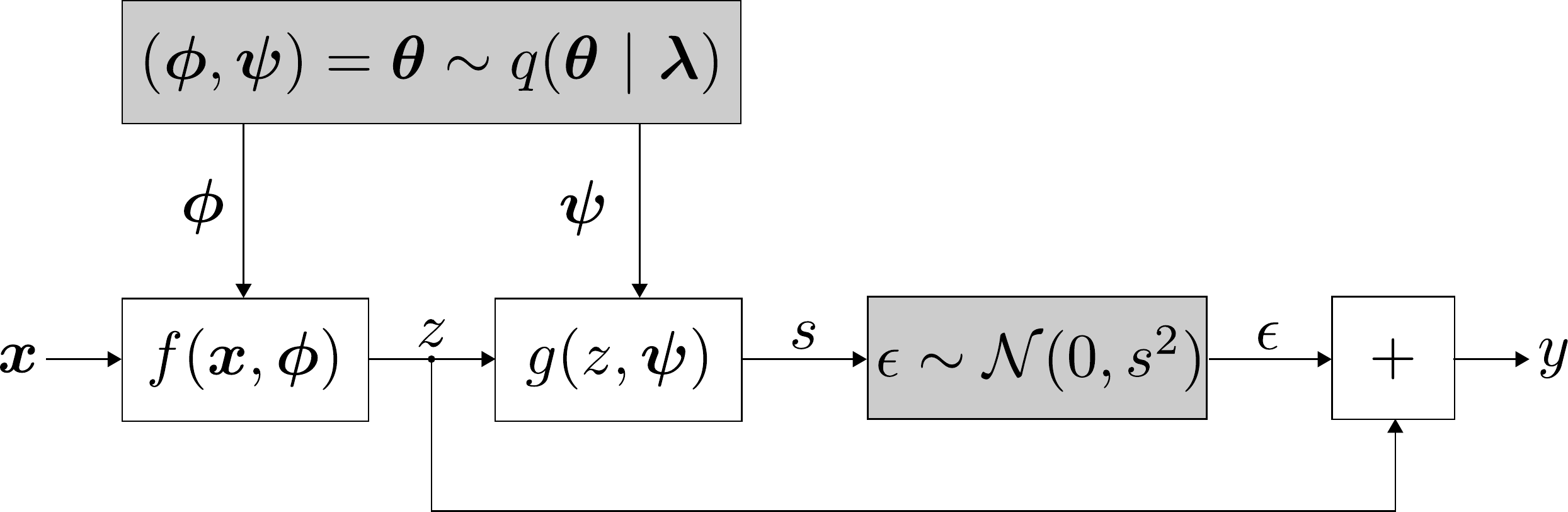}
\label{fig:full-arch}
}
\hfill
\subfloat[Composition of $f(\bm{x}, \bm{\phi})$]{
\includegraphics[width=1.0\textwidth]{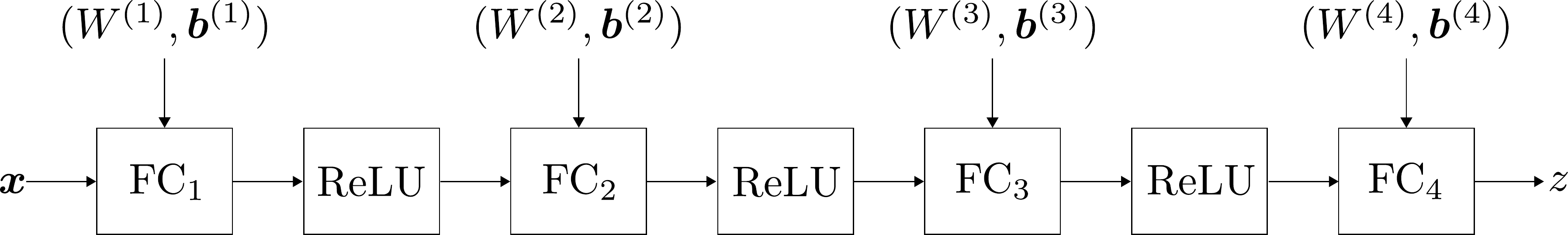}
\label{fig:arch-f}
}
\caption{The architecture of the BNNs used in this study is illustrated in (a). Probabilistic computations are colored grey. Variables $\bm{\phi}$ and $\bm{\psi}$ are drawn from the approximate posterior and used to compute the conditional mean flow rate, $f(\bm{x}, \bm{\phi})$, and noise standard deviation, $g(z, \bm{\psi})$. The composition of $f(\bm{x}, \bm{\phi})$ with four layers (three hidden) and $\bm{\phi} = \{(W^{(l)}, \bm{b}^{(l)})\}_{l=1}^{4}$ is shown in (b). Fully connected blocks perform the operation $\text{FC}_l(\bm{x}) = W^{(l)} \bm{x} + \bm{b}^{(l)}$.}
\label{fig:bnn-architecture}
\end{figure}

A schematic representation of the Bayesian neural network is shown in Figure \ref{fig:bnn-architecture}. The network architecture was fixed to three hidden layers, each with 50 nodes to which we apply the ReLU activation function \cite{Glorot2011}. Using practical recommendations in \cite{Bengio2012}, the network architecture may be large as long as regularization is used to prevent overfitting. The Adam optimizer \cite{Kingma2015b} with the learning rate set to 0.001 was used to train all networks. Early stopping with a validation dataset was used to determine an appropriate number of epochs to train the models to avoid overfitting \citep{Goodfellow2016}. The hyper-parameters were chosen by experimentation and using best practices. The models were implemented and trained using PyTorch \cite{Pytorch2019}.


\subsection{Prediction performance on historical data}
\label{sec:case-study-pred-historical}
To examine the predictive performance on historical data, a three months long period of contiguous data located in the middle of the dataset, when ordered chronologically, was set aside for testing. The rest of the data was used to train the models. During model development, a random sample of 20\% of the training data was used for model validation. The performance of each model type across the 60 wells was analyzed. Table \ref{tab:pred-hist-results} shows the $P_{10}$, $P_{25}$, $P_{50}$ (median), $P_{75}$, and $P_{90}$ percentiles of the MAPE across all wells. Detailed results which differentiate between test separator and MPFM measurements are reported in \ref{app:results}, Table \ref{tab:pred-hist-results-details}. 

\begin{table}[H]
\footnotesize
\caption{Prediction performance in terms of mean absolute percentage error on historical test data. The percentiles show the variation in performance among all wells.}
\begin{tabularx}{\textwidth}{ll*5{>{\raggedleft\arraybackslash}X}}
\toprule
Method and model & $P_{10}$ & $P_{25}$ & $P_{50}$ & $P_{75}$ & $P_{90}$\\
\midrule 
MAP-NN fixed homosc. & 1.8 & 2.8 & 5.1 & 8.3 & 16.0\\
VI-NN fixed homosc. & 1.4 & 2.6 & 4.8 & 8.5 & 12.8\\
VI-NN learned homosc.  & 1.3 & 2.4 & 5.3 &  8.4 & 13.3\\
VI-NN learned heterosc.  &  1.7 & 3.5 & 5.9 & 9.7 & 11.5\\
\bottomrule \noalign{\smallskip}
\end{tabularx}
\label{tab:pred-hist-results}
\end{table}

The results show that the four model types achieve similar performance to each other for the 75th and lower percentiles. The median MAPEs ($P_{50}$) lie in the range 4-6\%. A comparison of the 90th percentile performance indicates that models trained by variational inference are more robust in terms of modeling difficult wells. Regardless of the model type used, there are large variations in the performance on different wells, as seen by comparing the 10th and 90th percentiles. The best performing model achieved an error of 0.3\% for one of the wells. Yet, some models obtain an unsatisfactory large error. The overall worst-performing model (MAP-NN) achieved an error of 72.1\% for one of the wells. 

\begin{figure}[H]
\centering
\includegraphics[width=1.0\linewidth]{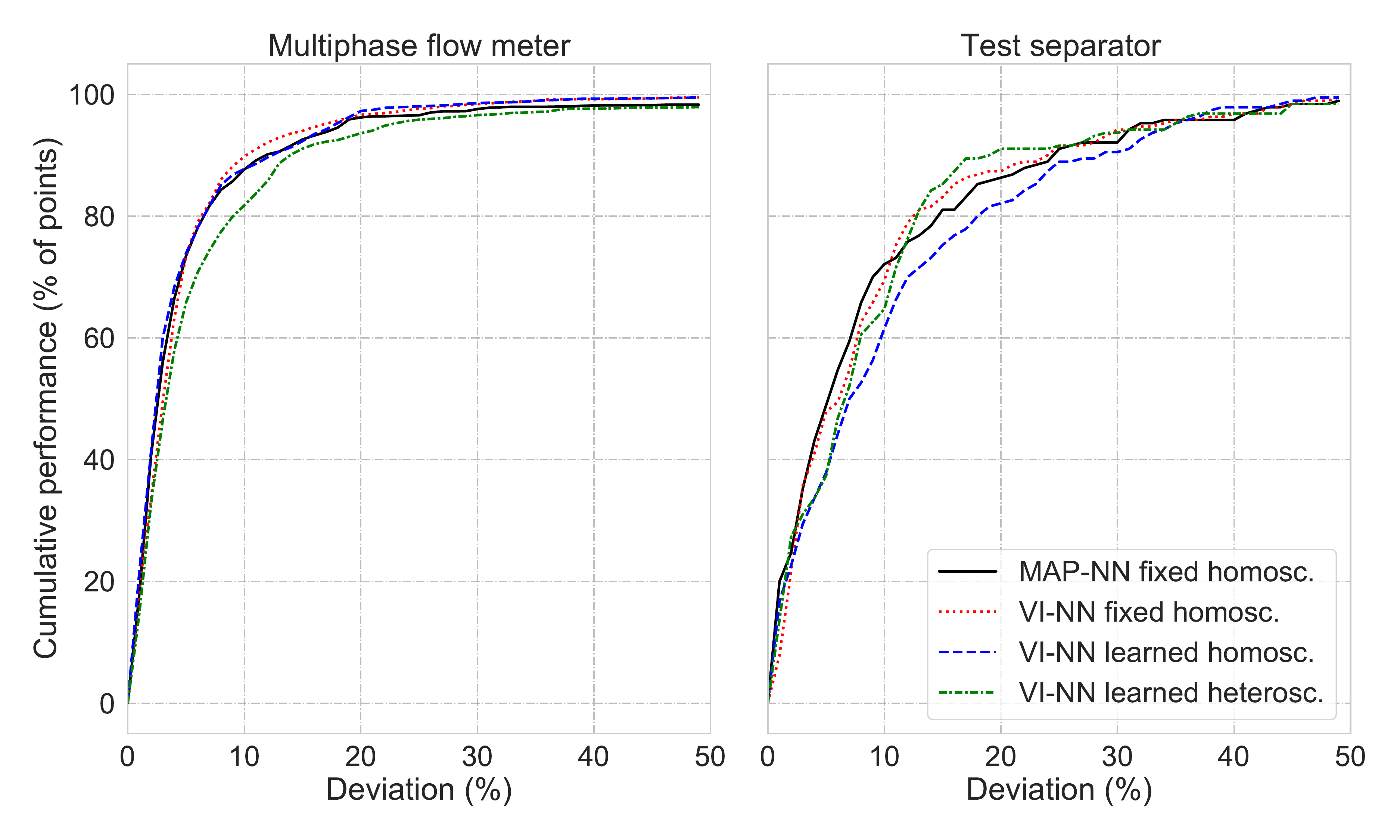}
\caption{Cumulative performance of the four models on historical test data. The cumulative performance is shown for wells with (left) MPFM and (right) test separator measurements.}
\label{fig:cum-perf-plot-historical}
\end{figure}

The cumulative performance of the four models is plotted in Figure \ref{fig:cum-perf-plot-historical}. The cumulative performance plot shows the percentage of test points that fall within a certain percent deviation from the actual measurements \cite{Corneliussen2005}. The figure shows that the models perform better on wells with MPFM measurements than on wells with test separator measurements. Again, similar performance of the four model types is observed.

\subsection{Prediction performance on future data}
\label{sec:case-study-pred-future}
The last three months of measurements were used to test the predictive performance on future data. The rest of the data was used to train the models. During model development, a random sample of 20\% of the training data was used for model validation. Table \ref{tab:pred-future-results} shows the percentiles of the MAPE for the different models on all 60 wells. Detailed results which differentiate between MPFM and test separator measurements are given in \ref{app:results}, Figure \ref{tab:pred-future-results-details}. 

\begin{table}[H]
\footnotesize
\caption{Prediction performance in terms of mean absolute percentage error on future test data. The percentiles show the variation in performance among all wells.}
\begin{tabularx}{\textwidth}{ll*5{>{\raggedleft\arraybackslash}X}}
\toprule
Method and model & $P_{10}$ & $P_{25}$ & $P_{50}$ & $P_{75}$ & $P_{90}$\\
\midrule 
MAP-NN fixed homosc.    & 3.7 & 5.6 & 12.4 &  24.1 &  40.0\\
VI-NN fixed homosc.      & 4.0 & 5.6 & 9.6 & 18.2 & 29.3\\
VI-NN learned homosc.    & 4.0 & 6.0 & 8.9 & 22.5 & 32.5\\
VI-NN learned heterosc.  & 4.0 & 5.0 & 9.2 & 15.7 & 24.3 \\
\bottomrule \noalign{\smallskip}
\end{tabularx}
\label{tab:pred-future-results}
\end{table}

Similarly to the case with historical test data, the performance of the four model types is comparable for the 50th and lower percentiles. The median MAPEs ($P_{50}$) lie in the range 8-13\%. For all model types, the 25\% best-performing models achieved a MAPE of less than 6\%. The best performing model obtained a MAPE of 1.1\% on one of the wells. This is in line with some of the best reported results in the literature; see Section \ref{sec:intro-trad}. Nevertheless, for each model type there is a large variation in performance among wells. The overall worst performing model achieved a MAPE of 48.7\%.

Comparing the performance for either the 75th or 90th percentile again indicates that models trained by variational inference are more robust in terms of modeling difficult wells. In this regard, the heteroscedastic VI-NN performs particularly well compared to the other model types.

\begin{figure}[H]
\centering
\includegraphics[width=1.0\linewidth]{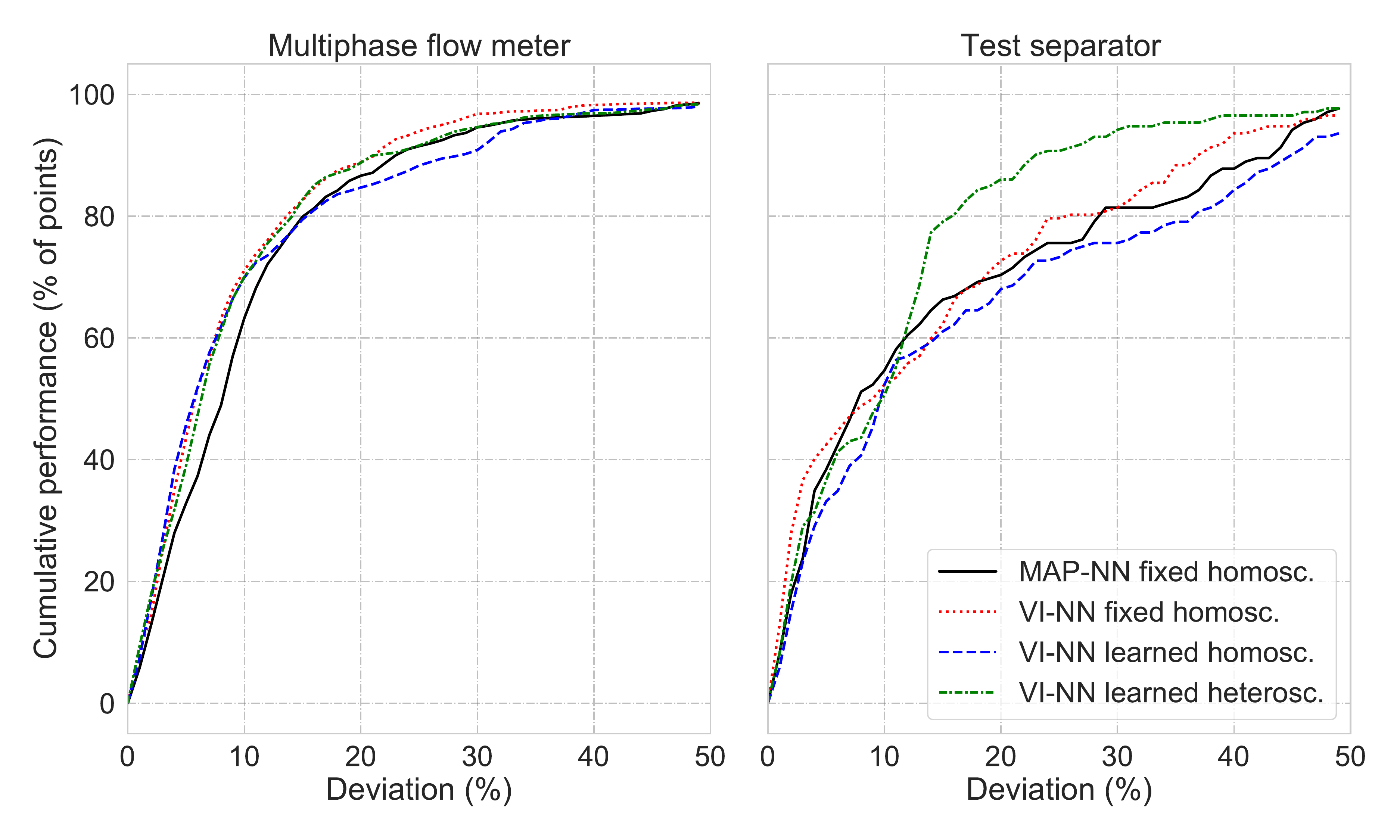}
\caption{Cumulative performance of the four models on future test data. The cumulative performance is shown for wells with (left) MPFM and (right) test separator measurements.}
\label{fig:cum-perf-plot-future}
\end{figure}

As seen from the cumulative performance plot in Figure \ref{fig:cum-perf-plot-future}, the four model types have similar performance to each other. The exception is the heteroscedastic VI-NN, which outperforms the other model types for wells with test separator measurements. As seen in the case of historical test data, the models perform better on wells with MPFM measurements than on well with test separator measurements. 

\subsection{Comparison of performance on historical and future data}
\begin{figure}[H]
\centering
\includegraphics[width=1.0\linewidth]{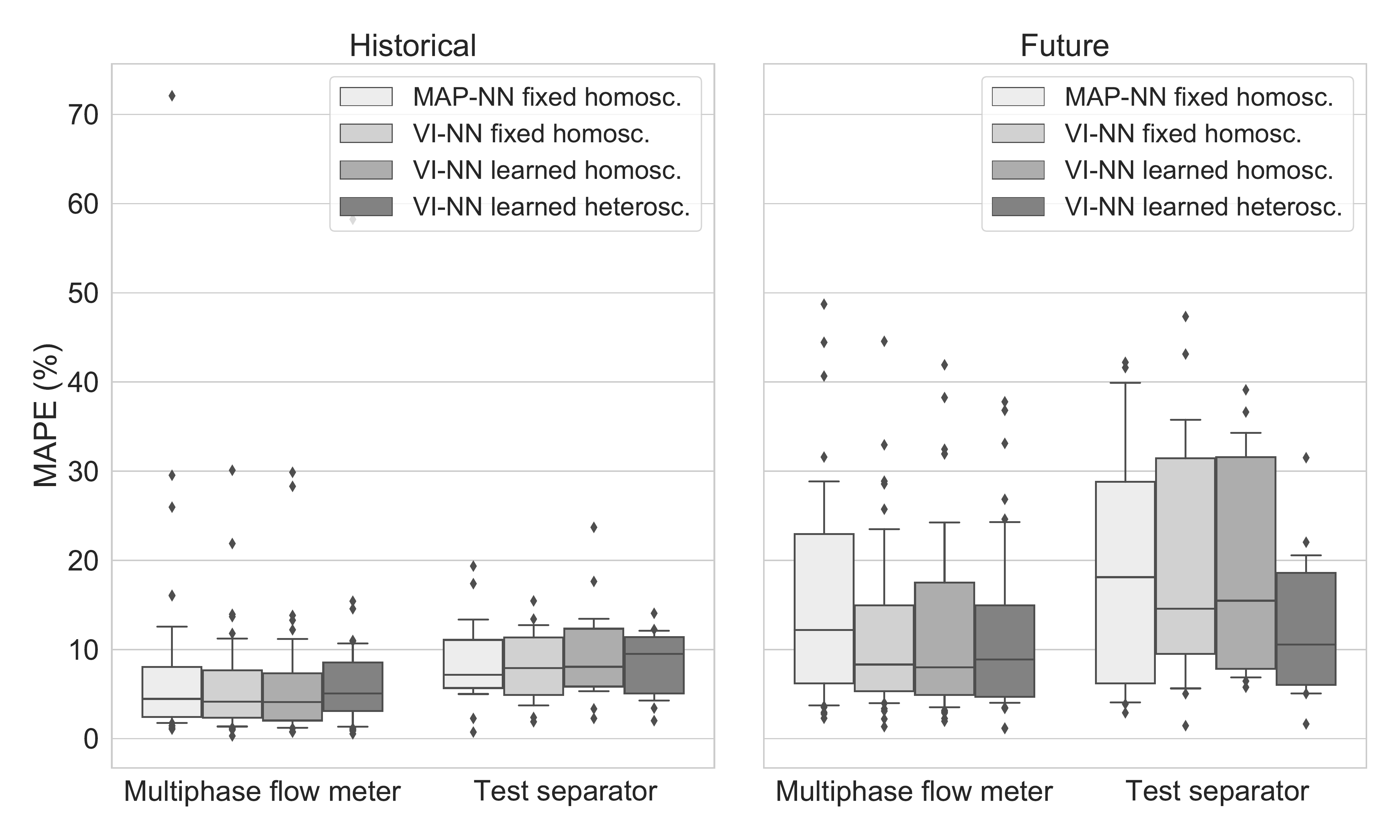}
\caption{Comparison of performance on historical and future data for the different models. The box plots differentiate between wells with multiphase flow meter and test separator measurements. The boxes show the $P_{25}$, $P_{50}$ (median), and $P_{75}$ percentiles. The whiskers show the $P_{10}$ and $P_{90}$ percentiles.}
\label{fig:mape-box-plot-hist-future}
\end{figure}

A comparison of the MAPEs on historical and future data is illustrated in Figure \ref{fig:mape-box-plot-hist-future}. The plots differentiate wells with MPFM and test separator measurements. In general, the prediction error is larger on future test data than on historical test data. There is also a larger variance in the performance on future test data. This indicates that it is harder to make predictions on future data, than on historical data. Further, observe that the errors are smaller for the wells with MPFM measurements than for the wells with test separator measurements in both the historical and future test data case. 

\subsection{Uncertainty quantification and analysis}
\label{sec:uncertainty-analysis-future}
\begin{figure}[ht!]
\centering
\subfloat[MPFM, historical]{
\includegraphics[width=0.23\textwidth]{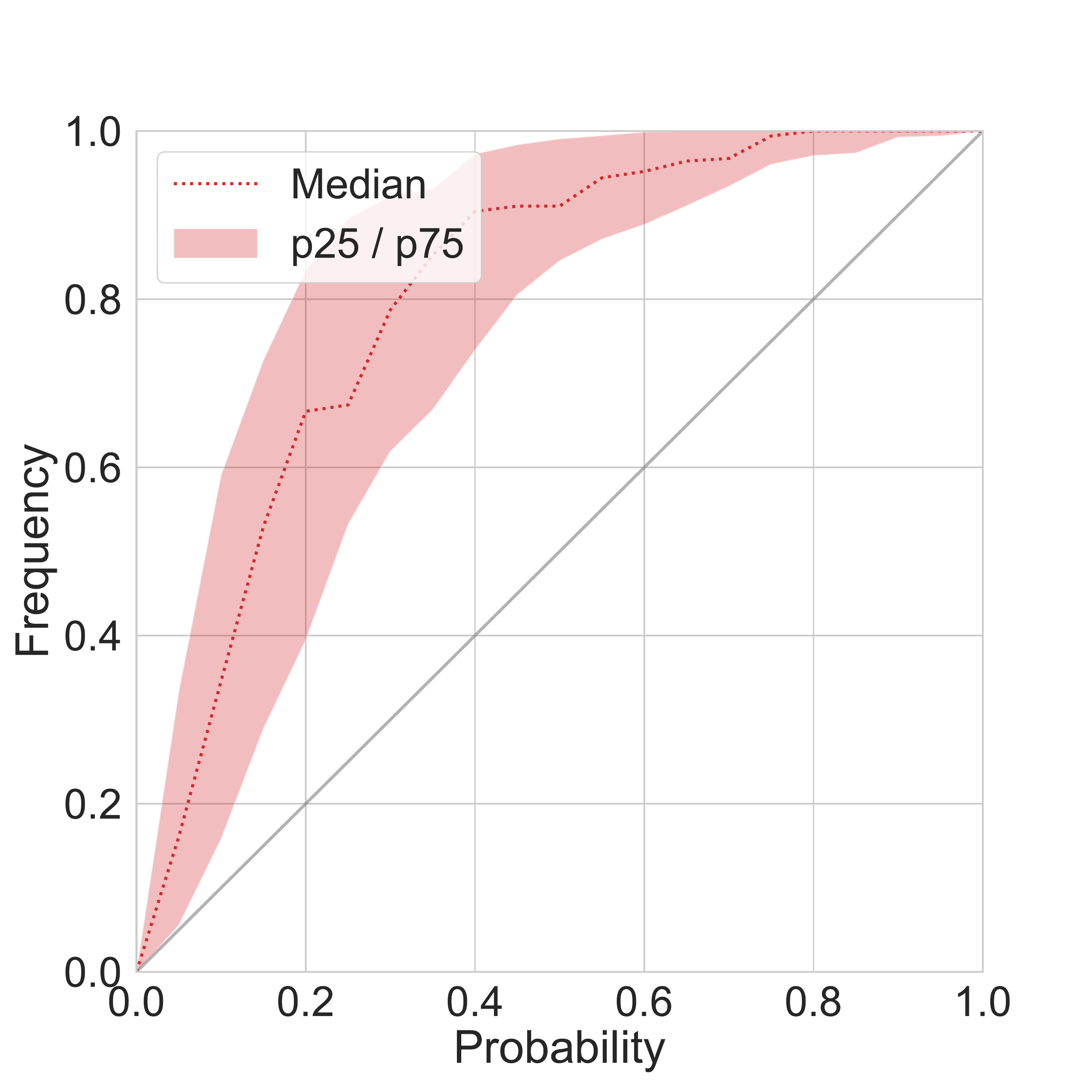}
\label{fig:pred-hist-calibration-fixed-mpfm}
}
\hfill
\subfloat[Separator, historical]{
\includegraphics[width=0.23\textwidth]{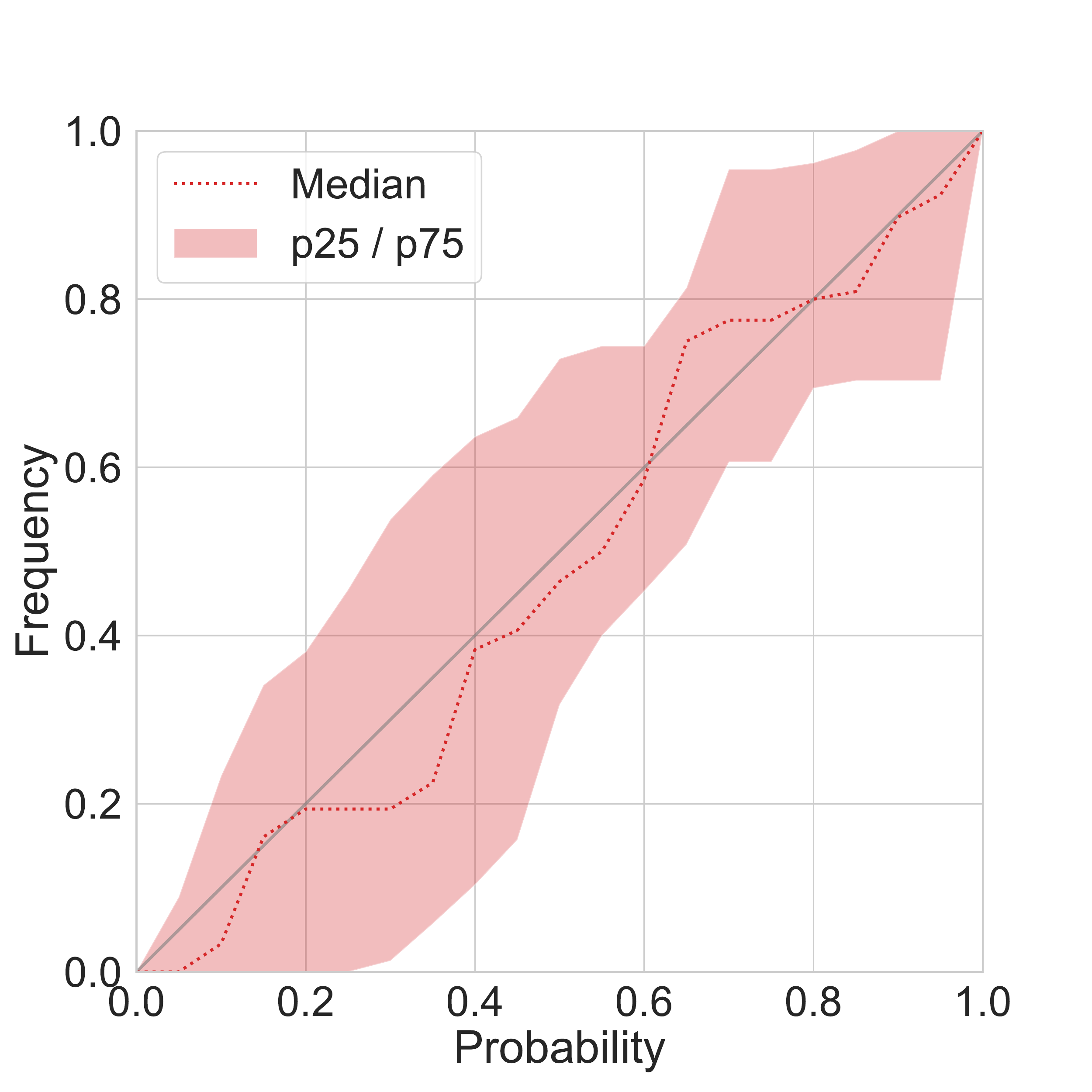}
\label{fig:pred-hist-calibration-fixed-septest}
}
\hfill
\subfloat[MPFM, future]{
\includegraphics[width=0.23\textwidth]{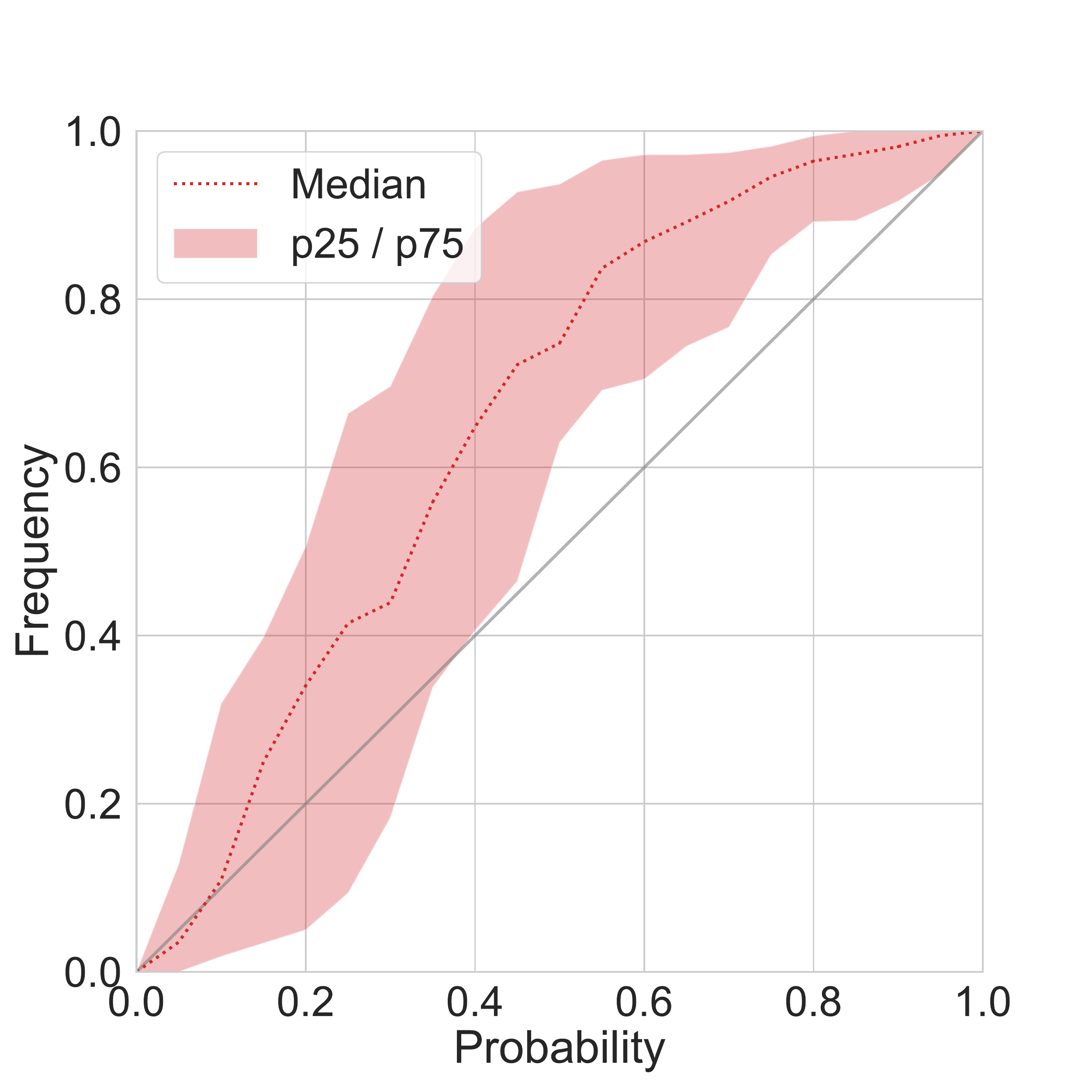}
\label{fig:calibration-plot-fixed-mpfm}
}
\hfill
\subfloat[Separator, future]{
\includegraphics[width=0.23\textwidth]{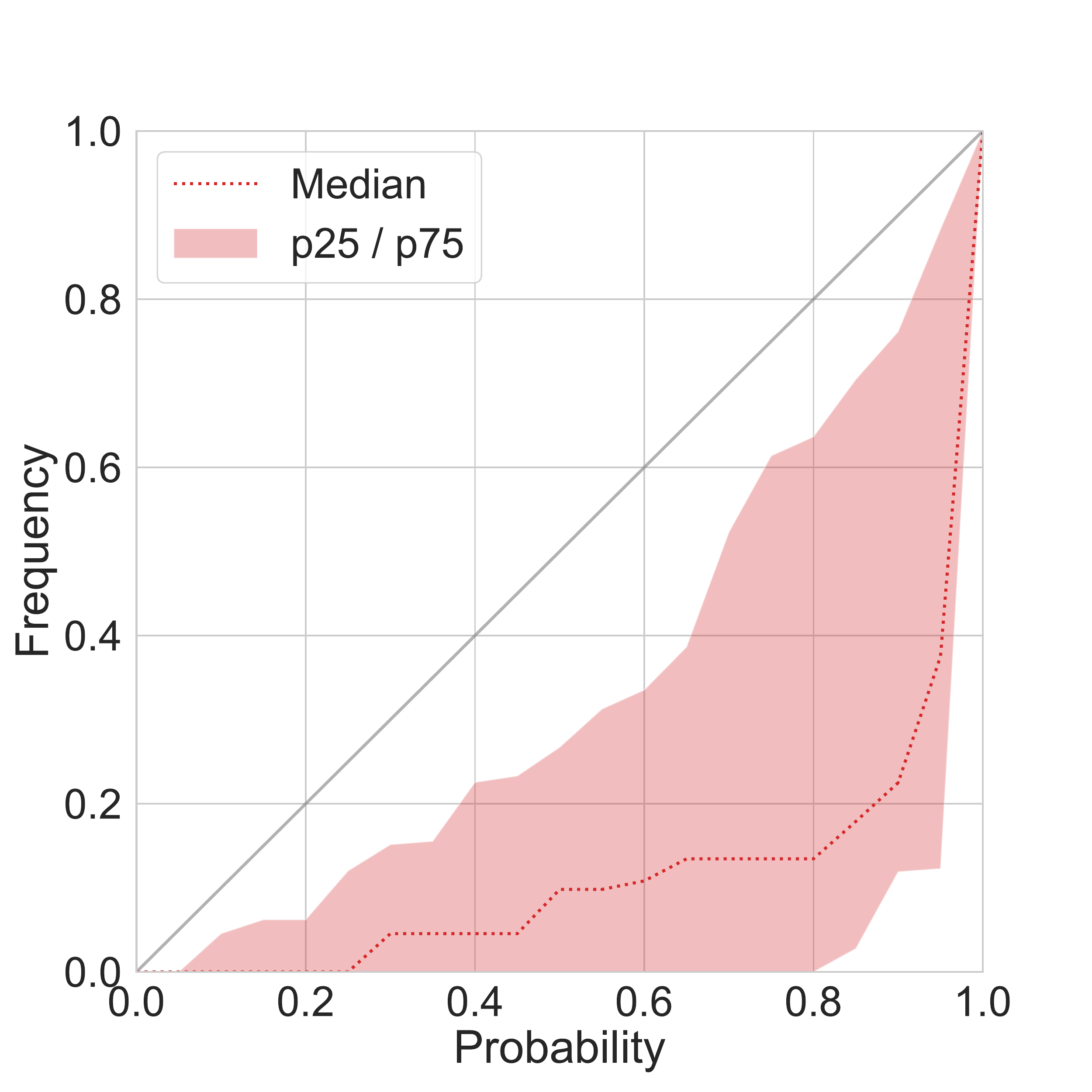}
\label{fig:calibration-plot-fixed-septest}
}
\vspace{1pt}
\subfloat[MPFM, historical]{
\includegraphics[width=0.23\textwidth]{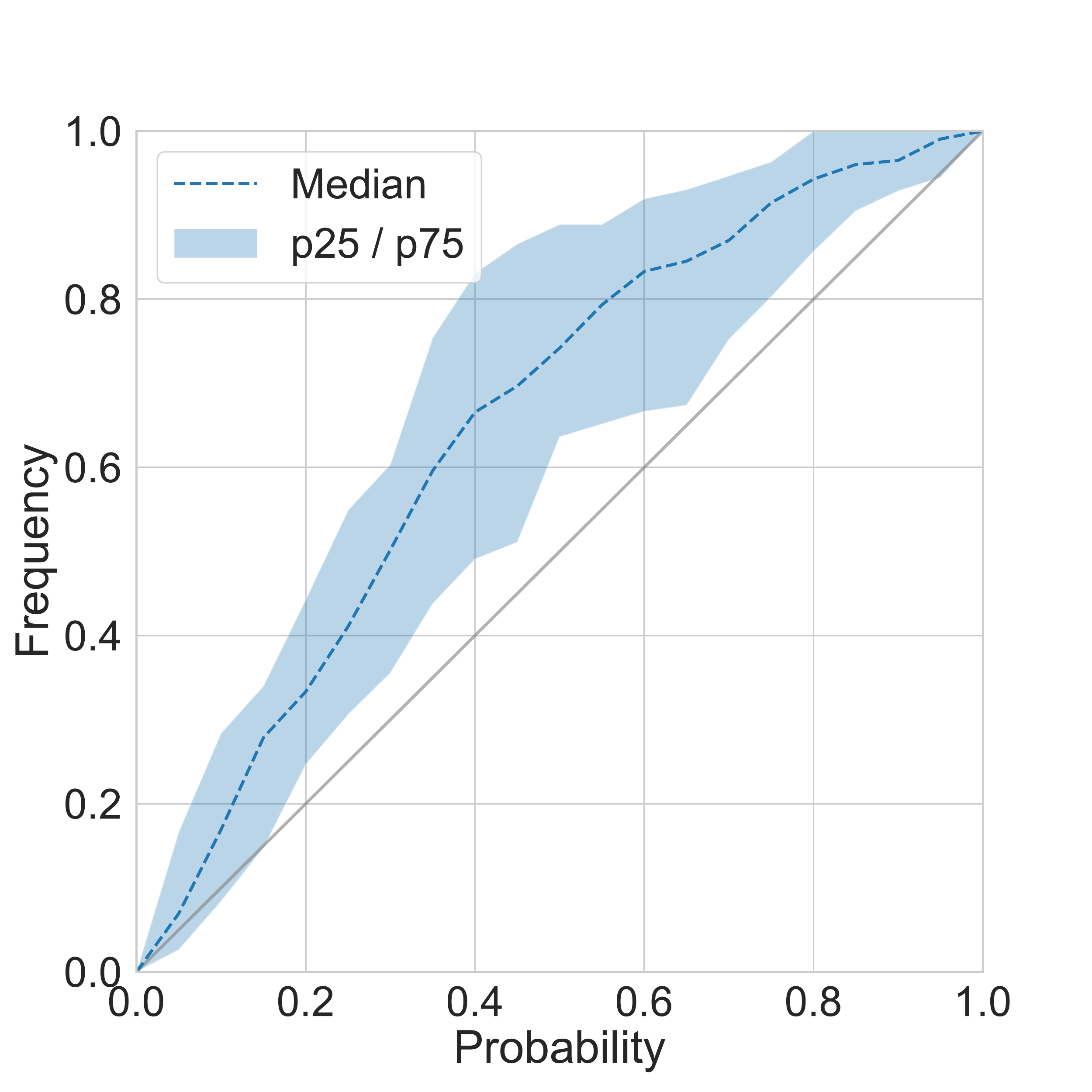}
\label{fig:pred-hist-calibration-homo-mpfm}
}
\hfill
\subfloat[Separator, historical]{
\includegraphics[width=0.23\textwidth]{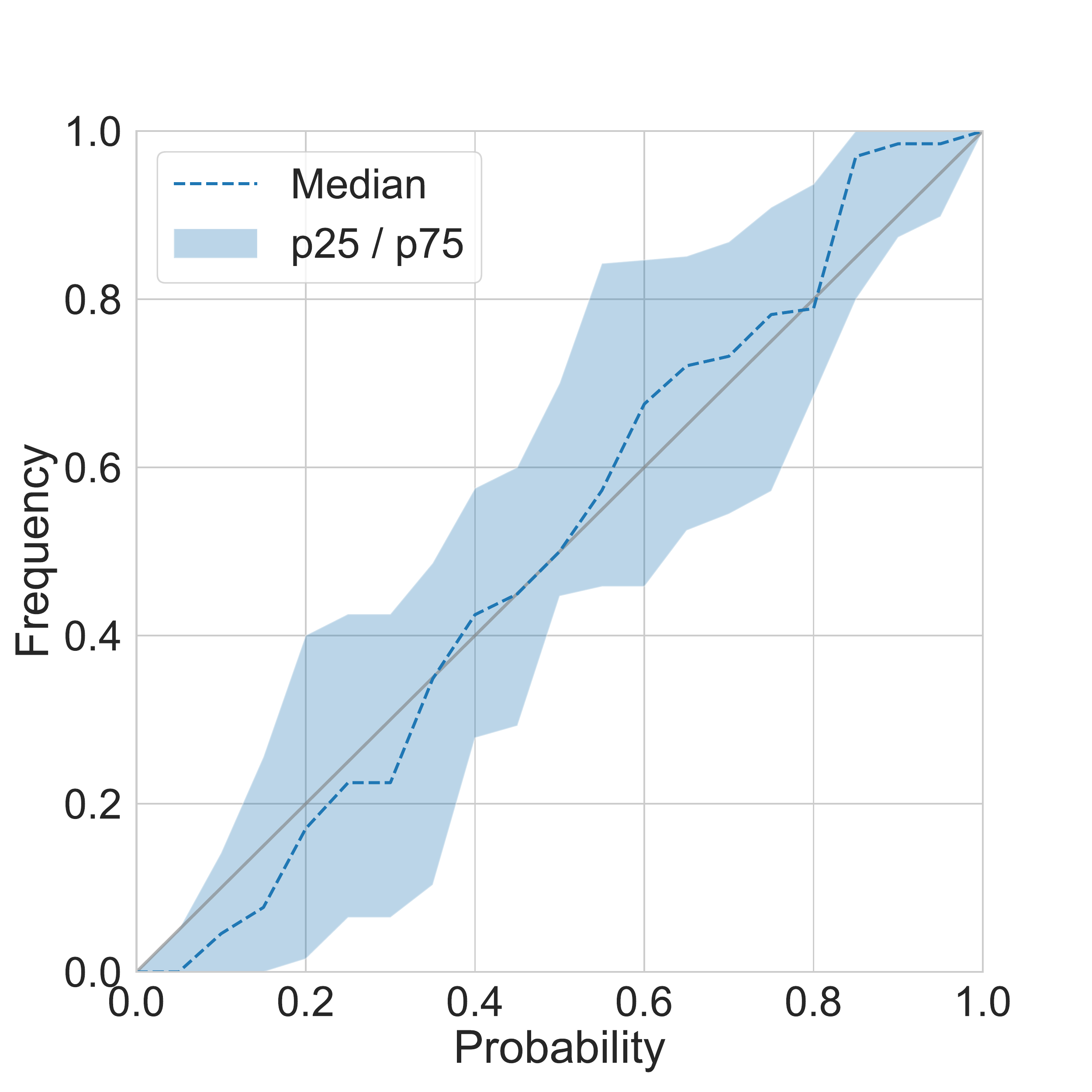}
\label{fig:pred-hist-calibration-homo-septest}
}
\hfill
\subfloat[MPFM, future]{
\includegraphics[width=0.23\textwidth]{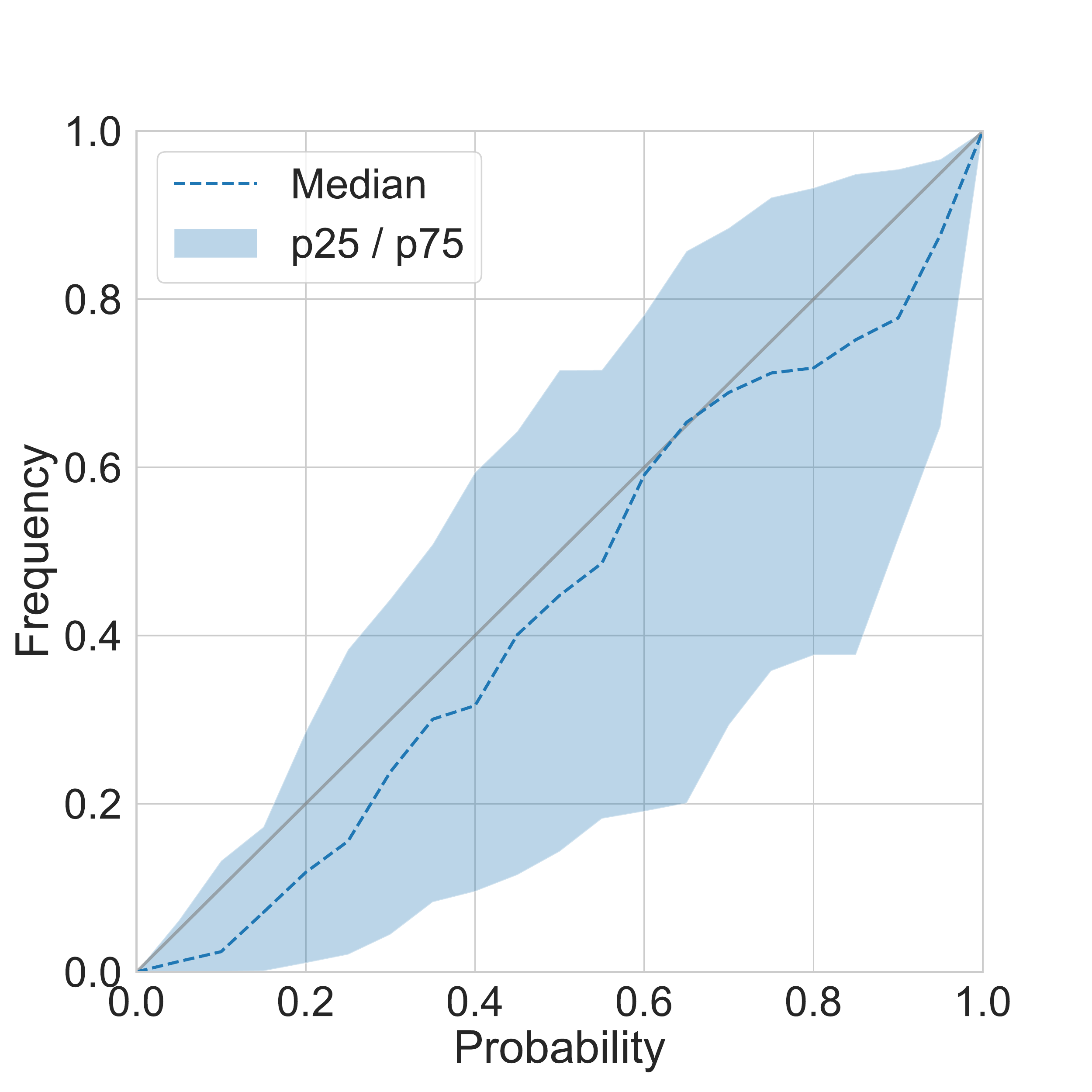}
\label{fig:calibration-plot-homo-mpfm}
}
\hfill
\subfloat[Separator, future]{
\includegraphics[width=0.23\textwidth]{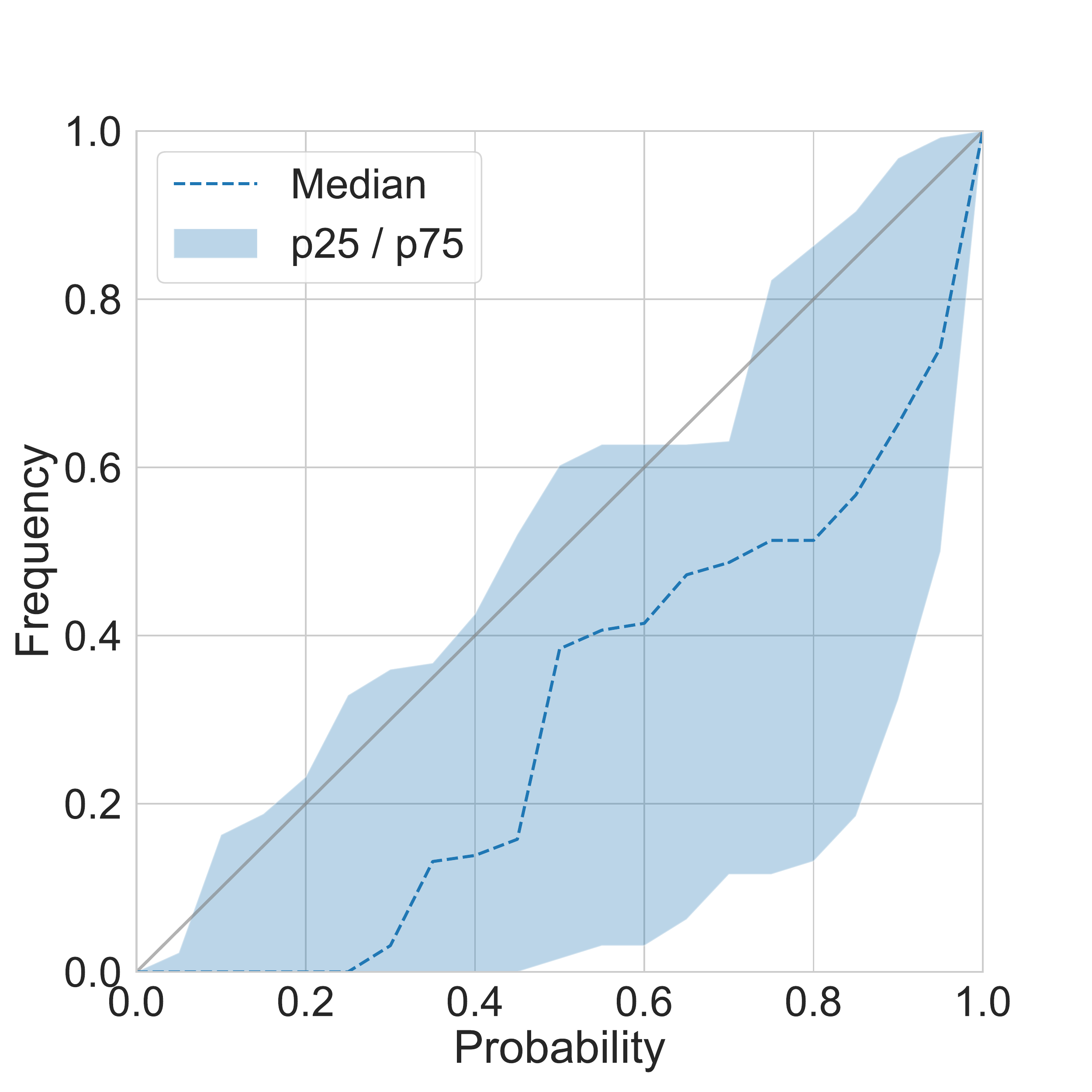}
\label{fig:calibration-plot-homo-septest}
}
\vspace{1pt}
\subfloat[MPFM, historical]{
\includegraphics[width=0.23\textwidth]{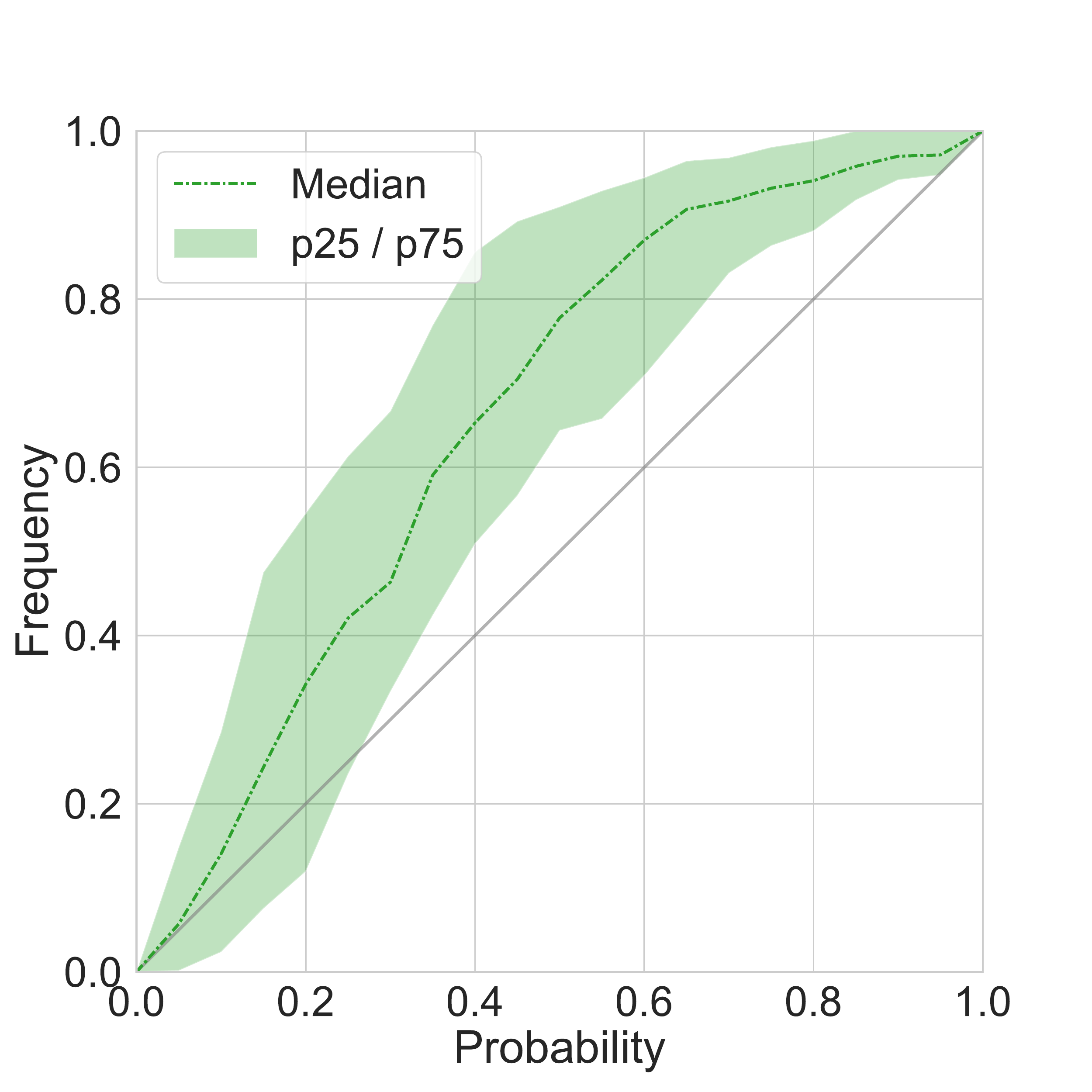}
\label{fig:pred-hist-calibration-hetero-mpfm}
}
\hfill
\subfloat[Separator, historical]{
\includegraphics[width=0.23\textwidth]{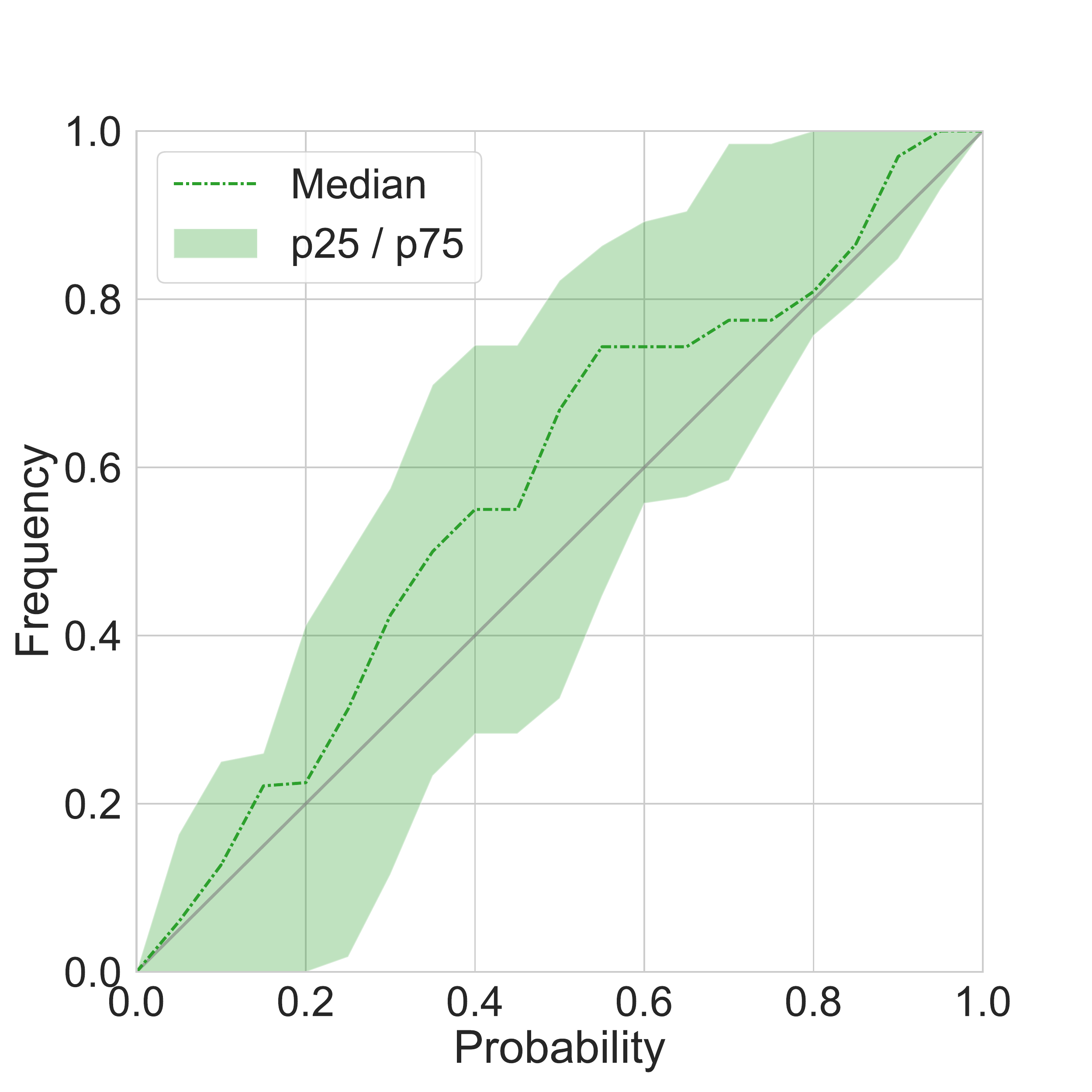}
\label{fig:pred-hist-calibration-hetero-septest}
}
\subfloat[MPFM, future]{
\includegraphics[width=0.23\textwidth]{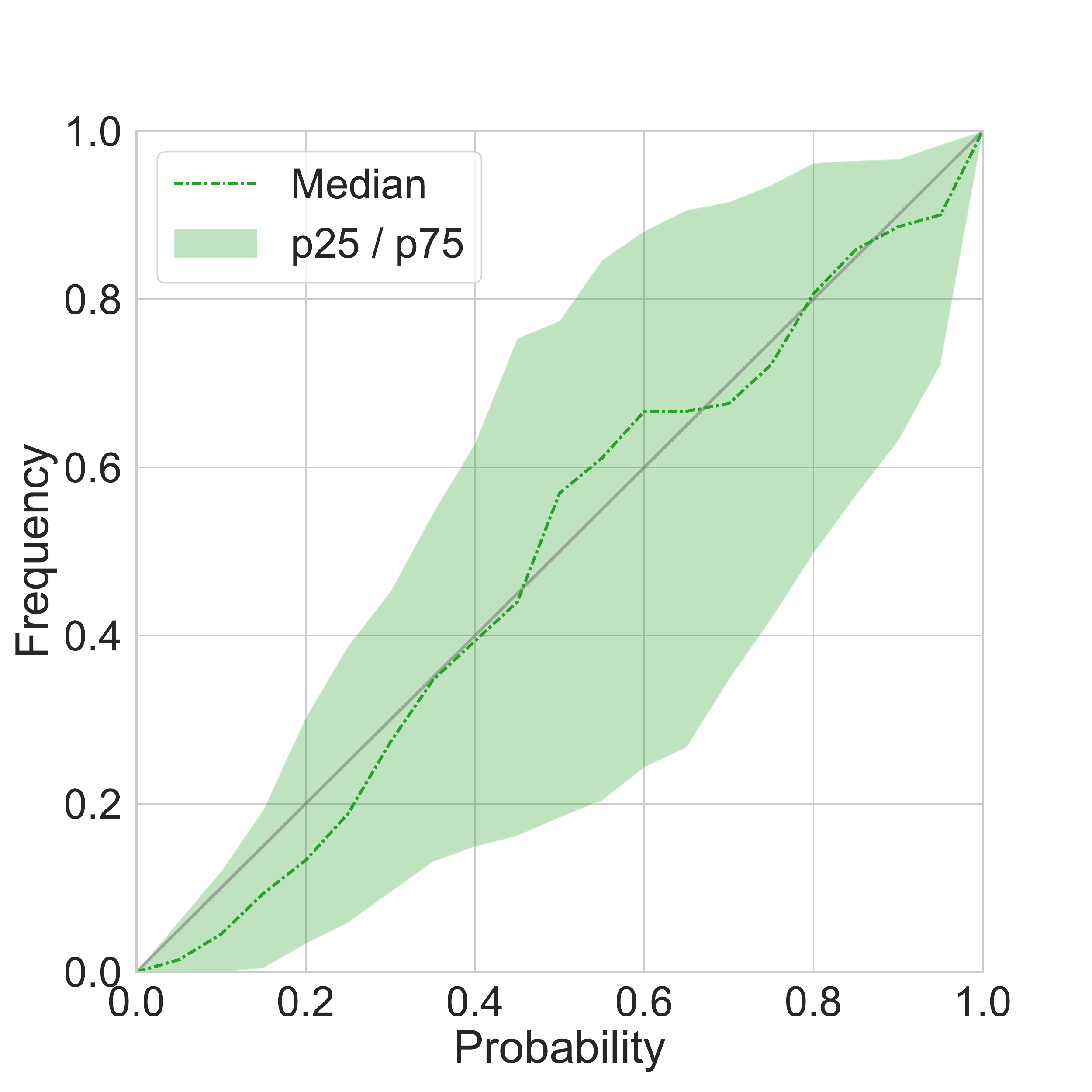}
\label{fig:calibration-plot-hetero-mpfm}
}
\hfill
\subfloat[Separator, future]{
\includegraphics[width=0.23\textwidth]{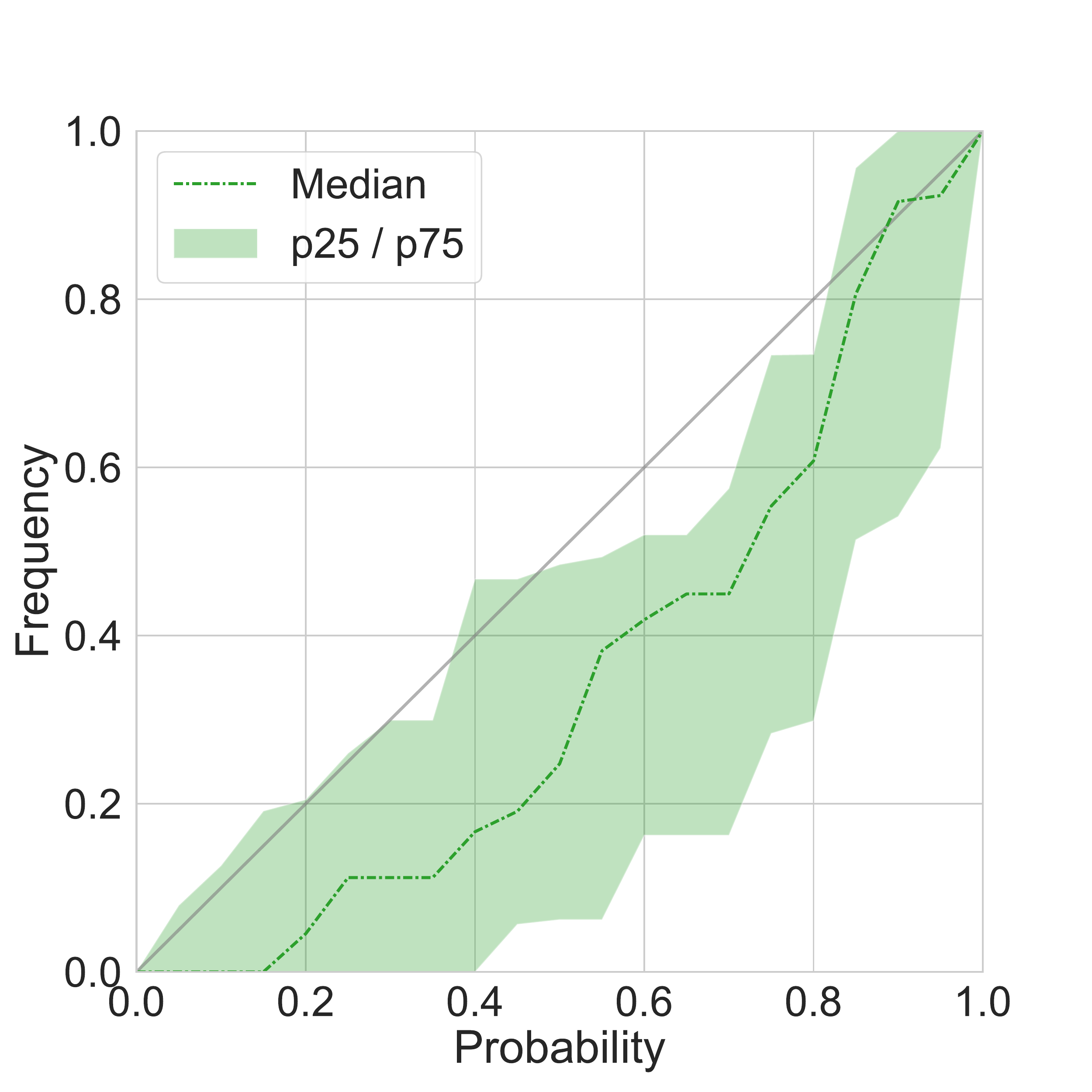}
\label{fig:calibration-plot-hetero-septest}
}
\caption{Calibration plots for fixed homoscedastic noise (a-d), learned homoscedastic noise (e-h), and learned heteroscedastic noise (i-l). Wells are grouped by measurement device, multiphase flow meter or test separator, and the calibration on historical test data (Section \ref{sec:case-study-pred-historical}) and future test data (Section \ref{sec:case-study-pred-future}) are shown. The median frequency is shown as a dashed line for each posterior interval (x-axis). The 25th and 75th percentiles (colored bands) show the variation in calibration across wells. A perfectly calibrated model would lie on the diagonal line $y=x$.}
\label{fig:calibration}
\end{figure}

In contrary to the MAP-NN models, the VI-NN models quantify the uncertainty in their predictions. To study the quality of the prediction uncertainty, we generated a calibration plot for the three different noise models using the test datasets from Section \ref{sec:case-study-pred-historical} and \ref{sec:case-study-pred-future}; see Figure \ref{fig:calibration}. The plot shows the frequency of residuals lying within varying posterior intervals. For instance, for a perfectly calibrated model, 20\% of the test points is expected to lie in the 20\% posterior interval centered about the posterior mean. In other words, the calibration curve of a perfectly calibrated model will lie on the diagonal gray line illustrated in the figures. The calibration of a model may vary across wells. To visualize the variance in model calibration, we have illustrated the (point-wise) 25th and 75th percentiles of the calibration curves obtained across wells. 

On historical data, the models trained on test separator measurements seem to be best calibrated. The models trained on MPFM measurements overestimate the uncertainty in their predictions. On future data, the results are reversed. The models trained on MPFM measurements are better calibrated and the models trained on test separator measurements all underestimate the prediction uncertainty. Overall, the calibration improves when the noise model is learned. This is seen clearly when comparing the fixed homoscedastic noise to the learned heteroscedastic noise model. The results are summarized in Table \ref{tab:BNN-coverage-probability}, which shows the coverage probabilities for the 95\% posterior interval (using the point-wise median in the calibration plots). 

\begin{table}[ht]
\footnotesize
\caption{Coverage probability (95\%)}
\begin{tabularx}{\textwidth}{ll*2{>{\raggedleft\arraybackslash}X}}
\toprule
Case & Method and model & Test sep. (\%) & MPFM (\%) \\
\midrule 
Future prediction & VI-NN fixed homosc. & 37.5 & 99.5 \\
& VI-NN learned homosc. & 81.0 & 87.7 \\
& VI-NN learned heterosc. & 92.3 & 90.0 \\
\multicolumn{4}{l}{} \\
Historical prediction & VI-NN fixed homosc. &  92.4 & 100.0 \\
& VI-NN learned homosc. & 98.5 & 99.1 \\
& VI-NN learned heterosc. & 100.0 & 97.2 \\
\bottomrule \noalign{\smallskip}
\end{tabularx}
\label{tab:BNN-coverage-probability}
\end{table}

\subsection{Effect of training set size on prediction performance}
\label{sec:case-study-dataset-set-size}
\begin{figure}[ht!]
\centering
\includegraphics[width=1.0\linewidth]{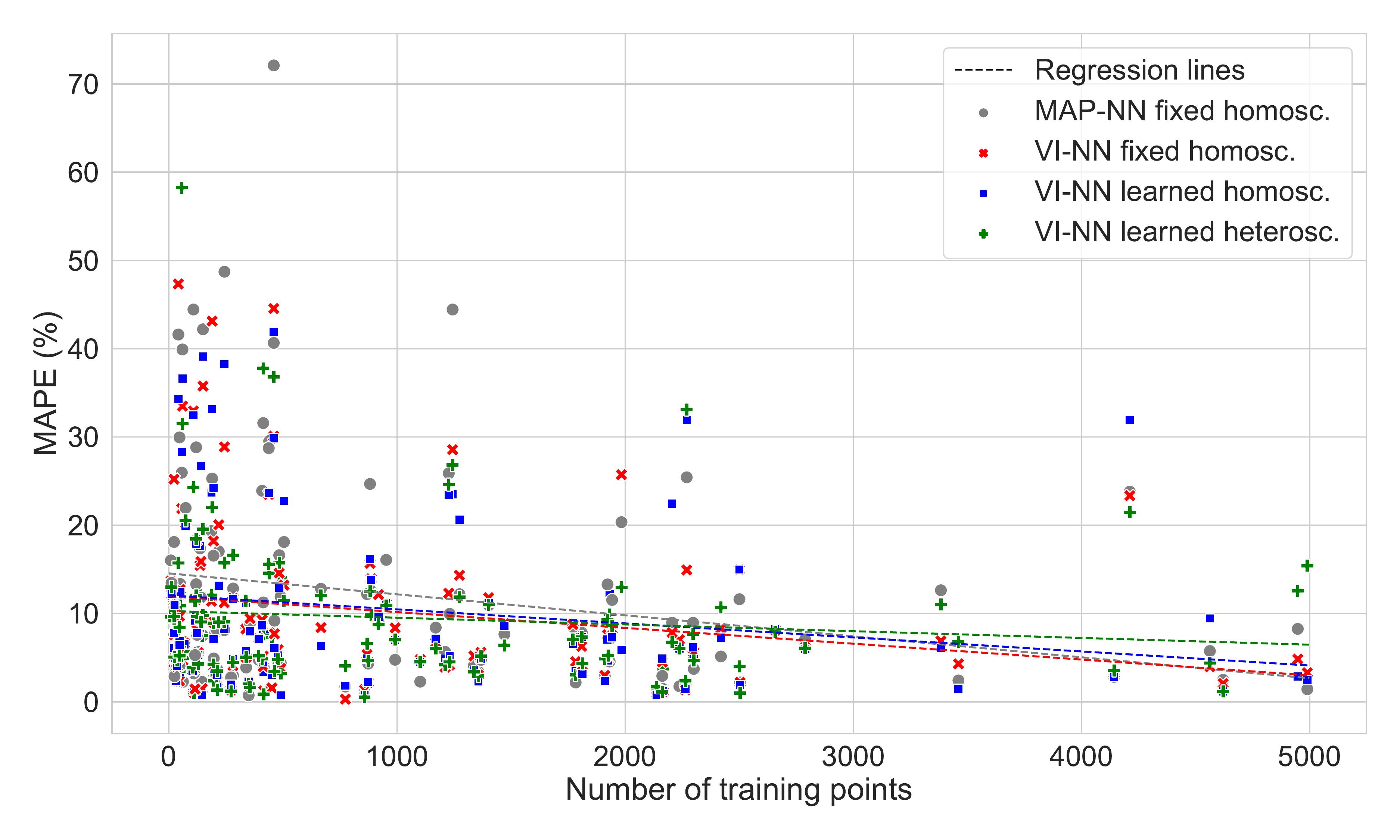}
\caption{The plot shows the mean absolute percentage error of the four models on historical and future test data for all wells. A regression line for each model shows the tendency of the error as the number of training points varies.}
\label{fig:mape-vs-ntrain}
\end{figure}

When analyzing the prediction performance of the four model types in Section \ref{sec:case-study-pred-historical} and \ref{sec:case-study-pred-future}, it was noticed that the prediction error tended to decrease as the training set size increased. This is illustrated in Figure \ref{fig:mape-vs-ntrain}, which shows the MAPEs for the different models and corresponding regression lines with negative slopes. This tendency is generally expected of machine learning models. On the other hand, previous studies such as \cite{AlQutami2018}, indicate that model performance does not necessarily improve when including data that is several years old. To closer inspect this effect, we compared models developed on successively larger training sets. 

To allow for an interesting range of dataset sizes a subset of 21 wells with 1200 or more MPFM measurements was considered. In a number of trials, a well from the subset and an instant of time at which to split the dataset into a training and test set, were randomly picked. Keeping the test set fixed, a sequence of training sets of increasing size was generated. The training sets were extended backwards in time with data preceding the test data. The following training set sizes were considered: 150, 200, 300, $\ldots$, 1100, where the increment is 100 between 300 and 1100. A MAP-NN model was developed for each of these training sets, using early stopping and validating against the last 100 data points. The test set size was also set to 100 data points, spanning on average 90 days of production. 

\begin{figure}[H]
\centering
\includegraphics[width=0.6\linewidth]{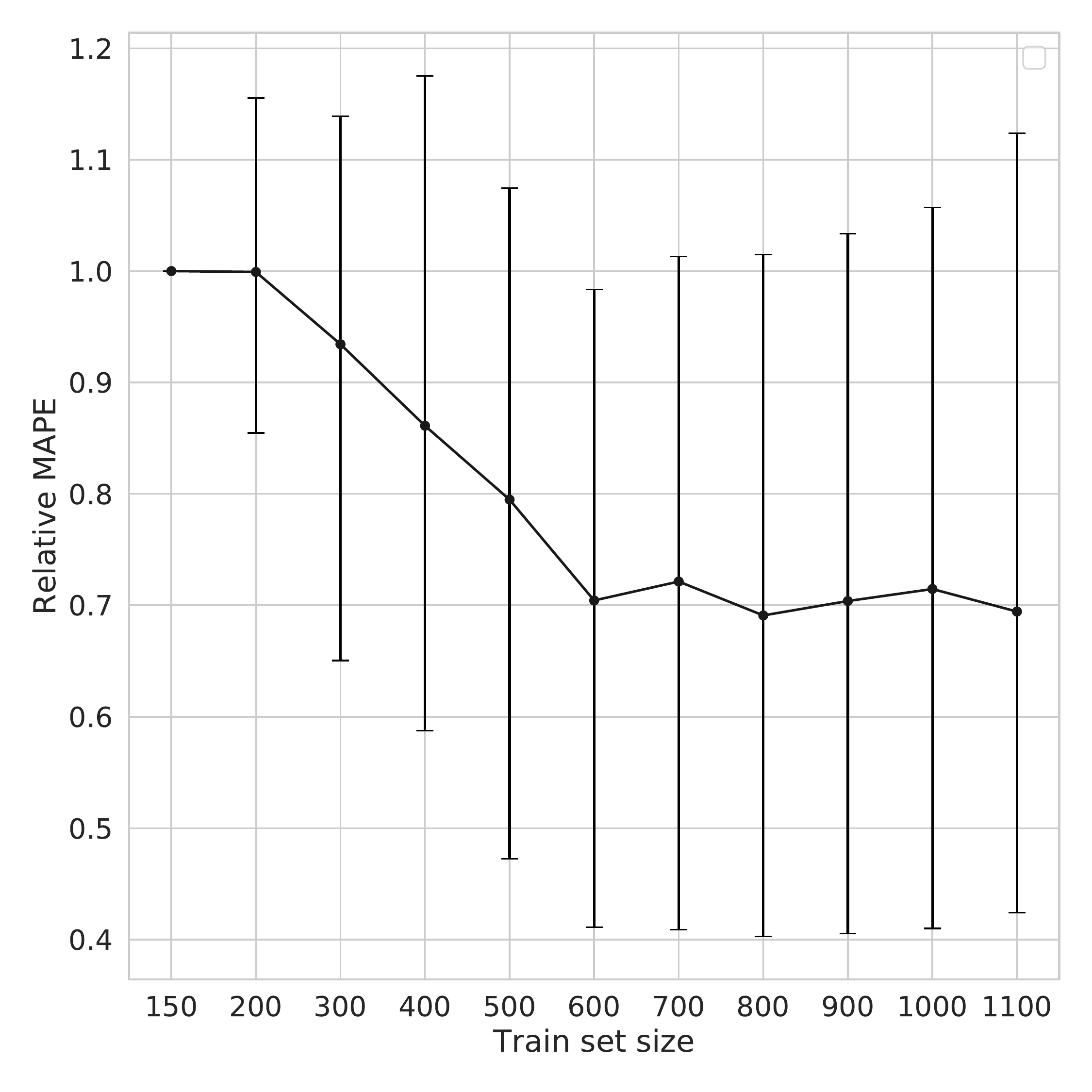}
\caption{Relative test errors of the MAP-NN model for increasing training set sizes. Shown are the medians and 50\% intervals of 400 trials.}
\label{fig:train-set-sizes}
\end{figure}

Denoting the test MAPE of the models by $E_{150}$, $E_{200}$, $E_{300}$, ..., $E_{1100}$, we computed relative MAPEs
\begin{equation}
    R_{k} = \frac{E_{k}}{E_{150}}, \text{ for } k \in \{150, 200, 300, \ldots, 1100\}.
\end{equation}
The relative errors indicate how the performance develops as the training set size increases, with a baseline at $R_{150} = 1$. The result of 400 trials is shown in Figure \ref{fig:train-set-sizes}.

\section{Discussion}
\label{sec:discussion}
In Section \ref{sec:introduction} some of the challenges faced by data-driven VFMs were discussed. These were: (1) low data volume, (2) low data variety (3) poor measurement quality, and (4) non-stationarity of the underlying process. Here we discuss the results in light of these challenges. All results are discussed in terms of MAPE values.

No widely used standard exists for VFM performance specification or requirements. Thus, the following performance requirements have been set by the authors to assess the commercial viability of a VFM: 1) predictive performance in terms of mean absolute percentage error on test data of 10\% or less, and 2) robustness in terms of achieving the above predictive performance for at least 90\% of wells. While these simple requirements lack a specification of the test data, we find them useful in the assessment of VFM performance. A VFM failing to meet these requirements would not be practical to use in industrial applications.

\subsection{Performance on historical and future test data}
First, we discuss the concern about the non-stationarity of the underlying process. This means the distribution of values seen during training is not necessarily the same as the distribution of values used for testing. The effect of this is best observed when comparing the performance on historical and future data, see Table \ref{tab:pred-hist-results} and \ref{tab:pred-future-results} and Figure \ref{fig:mape-box-plot-hist-future}. Looking at the upper and lower percentiles, we see the different models achieve performance in the range of 1-16\% error on historical data and 3-40\% error on future data. Since the strength of data-driven models lies with interpolation, rather than extrapolation, it is natural that the performance is worse on the future data case. Considering the VFM performance requirement of 10\% MAPE for 90\% of the wells, the performance is not acceptable for the historical or for the future data case. This indicates that the robustness of the models is inadequate for use in a commercial VFM. For real time applications, frequent model updates are likely required to achieve the VFM performance requirement. This raises the technical challenge of implementing a data-driven modeling approach. 

The study on dataset size in Section \ref{sec:case-study-dataset-set-size} further explores the development of data distributions and the effect older data has on future prediction errors. The result, seen in Figure \ref{fig:train-set-sizes}, indicates that additional data is only valuable up to a certain point, after which older data will no longer be useful when predicting future values. The point where this happens will naturally vary between wells. For the wells included here, this happens at 600 data points on average, for which the additional data is approximately 18 months or longer into the past. Looking at Figure \ref{fig:mape-vs-ntrain}, we again see the trend that wells with more data perform better, but only up to a certain point. We remark that insufficient model capacity would have a similar effect on the performance. However, we find this to be unlikely in this case study due to the high capacity and low training errors of the neural networks used.

At this point we remark that, for two observations $D_1, D_2 \in \dataset$, we model conditional independence $(D_1 \indep D_2 \mid \bm{\theta})$. While the observations result from preprocessing measurement data in a way that removes transients and decorrelates observations, we cannot guarantee independence due to the non-stationary process. With dependant observations, the modeling assumption of conditional independence is not satisfied since the models lack temporal dependencies. This is also true for most, if not all published models for data-driven VFM. Models that include temporal dependencies may be better suited to learn from past data.

A second concern raised was related to small data regimes, both in terms of data volume and data variety. The results mentioned above also illustrate the effect of small data. Looking at Figure \ref{fig:mape-vs-ntrain}, higher variance in performance is seen among wells with less than 700 data points. This is concerning because many of the wells, in particular those with test separator measurements as their primary source of data, have very few data points. Based on the median MAPE values in Figure \ref{fig:mape-box-plot-hist-future}, also given in Table \ref{tab:pred-hist-results-details} and \ref{tab:pred-future-results-details}, models trained on MPFM data outperforms the models trained on test separator data. This indicates that data quantity may outweigh data quality in the small-data regime. The difference in performance is also evident in the cumulative performance plots, see Figure \ref{fig:cum-perf-plot-historical} and \ref{fig:cum-perf-plot-future}. 

The wells that lie in the top quarter of performance achieved MAPE values comparable to the earlier works discussed in Section \ref{sec:intro-trad}. However, this performance seems difficult to achieve for the full set of wells. The difficulty in generalizing a single model architecture to a broad set of wells is troublesome for the potential commercialization of data-driven VFM.

\subsection{Noise models}
The last concern raised was poor data quality. In particular uncertainty in flow rate measurements, and potential gross errors in MPFM measurements.

The three different noise models perform similarly in terms of MAPE, on both historical and future data. The only exception being the learned heteroscedastic noise model, which performed better than the others on historical and future test data case when judged by the 90th percentile. This is believed to be because the heteroscedastic error term gives the objective function some added robustness towards large errors.

From the calibration plots in Figure \ref{fig:calibration}, we see that learning the noise model improves the calibration. The calibration curves for models trained on MPFM data generally lie above the curves for models trained on test separator data, both for historical and future predictions. This means that models trained on MPFM measurements are less confident in their predictions, even though they are trained on more data. It was suspected that models trained on MPFM data would reflect the increased uncertainty present in these measurements, but this is difficult to observe from the results. It is worth noting that the MPFM models are tested on MPFM data, so any systematic errors present in the MPFM measurements themselves will not be detected. 

Because the models have potentially large prediction errors, especially for future data, it is desirable that the model can assess its performance. The coverage probabilities reported in Table \ref{tab:BNN-coverage-probability} give us some confidence in the uncertainty estimates for the learned noise models, especially for the historical cases. 

Neither the homoscedastic or heteroscedastic noise models in \eqref{eq:homo-noise-model} and \eqref{eq:hetero-noise-model}, respectively, can capture complex noise profiles that depend on the flow conditions $\bm{x}$. As most flow meters are specialized to accurately measure flow rates for certain compositions and flow regimes, this is a potential drawback of the models. We leave it to later works to address such limitations, but note that with few adjustments the flow model in \eqref{eq:flow-model} can accommodate heteroscedasticity of a rather general form.

\subsection{Bayesian neural networks}
As stated in Section \ref{sec:introduction}, setting the priors on the parameters in the model is not a trivial task. In several papers, the Kullback-Leibler divergence term of the ELBO loss in \eqref{eq:elbo} is down-weighted to improve model performance due to poor priors \cite{Wenzel2020}. This remains a research question, however, in Section \ref{sec:prior-neural-network} one way of approaching prior specification in BNNs is described. The difficulty of setting priors combined with small data sets may make it difficult to successfully train models of this complexity. Still, the results are reasonable in the historical data case, and the estimated uncertainty is still better than only relying on point estimates.

\section{Concluding remarks}\label{sec:conclusion}
MAP estimation and VI for a probabilistic, data-driven VFM was presented and explored in a case study with 60 wells. 
The models achieve acceptable performance on future test data for approximately half of the studied wells. It is observed that models trained on historical data lack robustness in a changing environment. Frequent model updates are therefore likely required, which pose a technical challenge in terms of VFM maintenance.

Of the presented data challenges, the non-stationary data distribution is the most concerning. It means that models must have decent extrapolating properties if they are to be used in real-time applications. This is inherently challenging for data-driven approaches, and limits the performance of all the models considered in this paper. Of the models explored here, VI provided more robust predictions than MAP estimation on future test data.

The BNN approach is promising due to its ability to provide uncertainty estimates. Among these models, the heteroscedastic model had the best performance, indicating that a heteroscedastic model can be advantageous for flow rate measurements. However, it is challenging to obtain well-calibrated models due to the difficulty of setting meaningful priors on neural network weights, and the fact that priors play a significant role in small data regimes. As a result, the uncertainty estimates provided by the BNNs should be used with caution.

\subsection{Recommendations for future research}
We would suggest future research on data-driven VFM to focus on ways to overcome the challenges related to small data and non-stationary data distributions. Advances on these problems are likely required to improve the robustness and extrapolation capabilities of models to be used in real-time applications. We believe promising avenues of research to be: i) hybrid data-driven, physics-based models that allows for stronger priors; ii) data-driven architectures that enables learning from more data, for instance by sharing parameters between well models; iii) online learning to enable frequent model updates; and iv) modeling of temporal dependencies, for example using sequence models, to better capture time-varying boundary conditions.

\section*{Acknowledgements}
This work was supported by Solution Seeker AS.

\appendix

\section{Derivations}
\subsection{Log-likelihood of the flow rate model}
\label{app:log-likelihood}
The log-likelihood of the flow model in \eqref{eq:flow-model} with parameters $\bm{\theta} = (\bm{\phi}, \bm{\psi})$ on a dataset $\dataset = (X, \bm{y}) = \{(\bm{x}_i, y_i)\}_{i=1}^{N}$ is given by
\begin{equation}
\begin{aligned}
\log p(\bm{y} \condbar X, \bm{\theta}) &= \sum\limits_{i=1}^{N} \log p(y_i \condbar \bm{x}_i, \bm{\theta}) \\
&= \sum\limits_{i=1}^{N} \log \mathcal{N}(y_i \condbar f(\bm{x}_i, \bm{\phi}), g(f(\bm{x}_i, \bm{\phi}), \bm{\psi})^2) \\
&= - \frac{N}{2} \log (2 \pi) - \sum\limits_{i=1}^{N} \log g(f(\bm{x}_i, \bm{\phi}), \bm{\psi}) - \frac{1}{2} \left( \frac{y_i - f(\bm{x}_i, \bm{\phi})}{g(f(\bm{x}_i, \bm{\phi}), \bm{\psi})} \right)^2.
\end{aligned}
\end{equation}

With a homoscedastic noise model $g(z, \bm{\psi}) = \sigma_n = \text{const.}$, the log-likelihood simplifies to:
\begin{equation}
\log p(\bm{y} \condbar X, \bm{\theta}) = - \frac{N}{2} \log (2 \pi \sigma_n^2) - \frac{1}{2 \sigma_n^2} \sum\limits_{i=1}^{N} \left( y_i - f(\bm{x}_i, \bm{\phi}) \right)^2.
\label{eq:likelihood-fixed-noise}
\end{equation}

\subsection{Kullback-Leibler divergence term, $\kullb{q(\bm{\theta} \condbar \bm{\lambda})}{p(\bm{\theta})}$}
\label{app:KL-term}
Let the approximation $q(\bm{\theta} \condbar \bm{\lambda})$ and prior $p(\bm{\theta})$ be mean-field normal distributions of the random variables $\bm{\theta} \in \reals^K$. Assume that the approximation is parameterized with $\bm{\lambda} = (\bm{\mu}, \bm{\rho})$, where $\bm{\mu}$ is the mean and $\bm{\sigma} = \log(1 + \exp(\bm{\rho}))$ is the standard deviation of $q$. Then, the Kullback-Leibler divergence is given as:
\begin{equation}
\begin{aligned}
& \kullb{q(\bm{\theta} \condbar \bm{\lambda})}{p(\bm{\theta})} = \expectationp[q]{\log q(\bm{\theta} \condbar \bm{\lambda}) - \log p(\bm{\theta})} \\
&= \expectationp[q]{\sum\limits_{i=1}^{K} \log q(\theta_i \condbar \lambda_i) - \log p(\theta_i)} \\
&= \frac{1}{2} \expectationp[q]{\sum\limits_{i=1}^{K} - \log (2\pi \sigma_i^2) - \left(\frac{\theta_i - \mu_i}{\sigma_i} \right)^2 + \log (2\pi \bar{\sigma}_i^2) + \left(\frac{\theta_i - \bar{\mu}_i}{\bar{\sigma}_i} \right)^2} \\
&= \frac{1}{2} \left[ \sum\limits_{i=1}^{K} -2 \log \frac{\sigma_i}{\bar{\sigma}_i} - \frac{1}{\sigma_i^2}\underbrace{\expectationp[q_i]{(\theta_i - \mu_i)^2}}_{= \sigma_i^2} + \frac{1}{\bar{\sigma}_i^2} \expectationp[q_i]{(\theta_i - \bar{\mu}_i)^2} \right] \\
&= \frac{1}{2} \sum\limits_{i=1}^{K} \left[-1 - 2\log \frac{\sigma_i}{\bar{\sigma}_i} + \frac{1}{\bar{\sigma}_i^2} \expectationp[q_i]{(\theta_i - \bar{\mu}_i)^2} \right] \\
&= \frac{1}{2} \sum\limits_{i=1}^{K} \left[-1 - 2\log \frac{\sigma_i}{\bar{\sigma}_i} + \left( \frac{\mu_i - \bar{\mu}_i}{\bar{\sigma}_i} \right)^2 + \left( \frac{\sigma_i}{\bar{\sigma}_i} \right)^2 \right]
\end{aligned}
\end{equation}

\section{Results}\label{app:results}
\begin{table}[H]
\footnotesize
\caption{Prediction performance on historical test data for each well group. Reported values are the $P_{10}$, $P_{25}$, $P_{50}$, $P_{75}$, and $P_{90}$ percentiles for the statistics root mean square error (RMSE) and mean absolute percentage error (MAPE).}
\begin{tabularx}{\textwidth}{llrr}
\toprule
Well group & Method and model & RMSE  & MAPE \%  \\
\midrule 
All & MAP-NN fixed homosc. & 0.4, 0.7, 1.1, 1.7, 3.0 & 1.8, 2.8, 5.1, 8.3, 16.0\\
& VI-NN fixed homosc. & 0.3, 0.5, 1.0, 2.1, 3.0 & 1.4, 2.6, 4.8, 8.5, 12.8\\
& VI-NN learned homosc. & 0.3, 0.5, 1.0, 2.0, 3.0 & 1.3, 2.4, 5.3,  8.4, 13.3\\
& VI-NN learned heterosc. & 0.4, 0.6, 1.2, 1.9, 3.0 &  1.7, 3.5, 5.9, 9.7, 11.5\\
\multicolumn{4}{l}{} \\
Test sep. & MAP-NN fixed homosc. & 0.4, 0.8, 1.5, 1.7, 3.0 & 3.1, 5.7, 7.2, 11.1, 16.2 \\
& VI-NN fixed homosc. & 0.5, 0.8, 1.6, 2.2, 4.3 & 2.8, 4.9, 7.9, 11.3, 13.2 \\
& VI-NN learned homosc. & 0.6, 1.1, 1.7, 2.1, 3.1 & 3.9, 5.8, 8.1, 12.3, 16.4 \\
& VI-NN learned heterosc. & 0.5, 1.0, 1.7, 2.1, 3.9  & 3.7, 5.1, 9.5, 11.4, 12.2\\
\multicolumn{4}{l}{} \\
MPFM & MAP-NN fixed homosc. &0.3, 0.6, 1.0, 1.6, 2.8 & 1.8, 2.4, 4.5,  8.1, 14.3 \\
& VI-NN fixed homosc. & 0.3, 0.4, 1.0, 1.9, 2.9 & 1.3, 2.3, 4.1, 7.7, 11.5\\
& VI-NN learned homosc. & 0.3, 0.4, 0.7, 1.6, 3.0  & 1.2, 2.0, 4.1, 7.3, 11.7 \\
& VI-NN learned heterosc. & 0.4, 0.5, 1.2, 1.5, 2.9 & 1.3, 3.1, 5.1, 8.6, 10.8 \\
\bottomrule \noalign{\smallskip}
\end{tabularx}
\label{tab:pred-hist-results-details}
\end{table}

\begin{table}[H]
\footnotesize
\caption{Prediction performance on future test data for each well group. Reported values are the $P_{10}$, $P_{25}$, $P_{50}$, $P_{75}$, and $P_{90}$ percentiles for the statistics root mean square error (RMSE) and mean absolute percentage error (MAPE).}
\begin{tabularx}{\textwidth}{llrr}
\toprule
Well group & Method and model & RMSE  & MAPE \%  \\
\midrule 
All & MAP-NN fixed homosc. & 0.8, 1.2, 2.1, 4.0, 6.1 & 3.7, 5.6, 12.4, 24.1, 40.0\\
& VI-NN fixed homosc. & 0.6, 1.1, 1.8, 3.5, 5.2 & 4.0, 5.6, 9.6, 18.2, 29.3\\
& VI-NN learned homosc. & 0.7, 1.2, 1.9, 3.3, 5.5 & 4.0, 6.0, 8.9, 22.5, 32.5\\
& VI-NN learned heterosc. & 0.6, 1.1, 1.7, 3.1, 4.5 & 4.0, 5.0, 9.2, 15.7, 24.3 \\
\multicolumn{4}{l}{} \\
Test sep. & MAP-NN fixed homosc. & 0.8, 1.0, 1.6, 3.0, 6.7 & 3.9, 6.2, 18.1, 28.8, 41.1  \\
& VI-NN fixed homosc. & 0.3, 1.0, 2.1, 3.2, 8.0 & 5.2, 9.5, 14.6, 31.4, 40.9 \\
& VI-NN learned homosc. & 0.6, 1.3, 1.9, 3.6, 5.9 & 6.6, 7.8, 15.5, 31.6, 35.9\\
& VI-NN learned heterosc. & 0.4, 1.2, 1.6, 2.3, 2.9 & 5.1, 6.0, 10.6, 18.6, 21.6\\
\multicolumn{4}{l}{} \\
MPFM & MAP-NN fixed homosc. & 0.9, 1.2, 2.4, 4.2, 5.7 & 3.7, 6.2, 12.2, 23.0, 30.2 \\
& VI-NN fixed homosc. & 0.8, 1.3, 1.8, 3.5, 4.6 & 4.0, 5.3, 8.3, 15.0, 24.6 \\
& VI-NN learned homosc. & 0.7, 1.1, 1.9, 3.1, 5.2 & 3.4, 4.9, 8.0, 17.5, 28.1 \\
& VI-NN learned heterosc. & 0.7, 1.0, 1.8, 3.3, 4.6 & 3.8, 4.7, 8.9, 14.9, 24.5 \\
\bottomrule \noalign{\smallskip}
\end{tabularx}
\label{tab:pred-future-results-details}
\end{table}

\bibliography{references}

\end{document}